\documentclass[11pt]{article}
\usepackage[preprint]{acl}
\usepackage{times}
\usepackage{latexsym}
\usepackage[T1]{fontenc}
\usepackage[utf8]{inputenc}
\usepackage{microtype}
\usepackage{xurl}
\usepackage{inconsolata}
\usepackage{graphicx}
\usepackage{amsmath,amssymb}
\usepackage{booktabs,array,makecell,tabularx,adjustbox}
\usepackage{enumitem}
\usepackage{placeins}
\usepackage{needspace}
\usepackage{stfloats}
\usepackage[table]{xcolor}
\usepackage{listings}
\usepackage[most]{tcolorbox}
\tcbuselibrary{breakable,skins,listings}
\usepackage[ruled,vlined]{algorithm2e}
\usepackage{float}
\SetKwComment{Comment}{/* }{ */}

\definecolor{signalgreen}{RGB}{0,128,0}
\definecolor{noisered}{RGB}{200,50,50}
\definecolor{neutralgray}{RGB}{100,100,100}
\definecolor{macaronred}{RGB}{244,166,166}
\definecolor{macaronblue}{RGB}{159,202,236}

\newcommand{\ret}[1]{\textcolor{neutralgray}{\footnotesize #1}}

\newcommand{\stepp}[1]{\textcolor{black}{\small #1}}
\newcommand{\good}[1]{\textcolor{signalgreen}{\footnotesize #1}}
\newcommand{\bad}[1]{\textcolor{noisered}{\footnotesize #1}}
\lstdefinestyle{trace}{
  basicstyle=\footnotesize\ttfamily,numbers=none,numbersep=0pt,
  breaklines=true,postbreak=\mbox{\textcolor{gray}{$\hookrightarrow$}\space},
  showstringspaces=false,columns=flexible,keepspaces=true,
  xleftmargin=0pt,framexleftmargin=0pt,aboveskip=0pt,belowskip=0pt,
  escapeinside={(*@}{@*)}}
\lstnewenvironment{tracelist}{
  \lstset{style=trace,numbers=none,numbersep=0pt,xleftmargin=0pt,
    framexleftmargin=0pt,breaklines=true,linewidth=\linewidth}}{}
\definecolor{findingBg}{RGB}{232,236,243}
\definecolor{findingRule}{RGB}{60,60,60}

\newcommand{\benchname}{OmniStepBench}
\title{OmniStepBench: Deployment-Aware Agent Evaluation across Modalities and Task Complexity}
\author{
  \bfseries Yu Liu$^{1,2,3*\dagger}$, Zhilin Liu$^{3*}$,
  Zhiwei Yang$^{1,2,3*}$, Shaojie Zhang$^{3*}$, Zheyuan Deng$^{4}$ \\
  \bfseries Tingwei Huang$^{3}$, Zhenbo Luo$^{3}$, Lei Jiang$^{1,2}$,
  Yanbing Liu$^{2}$, Pei Fu$^{3\dagger}$ \\
  \normalfont $^{1}$Institute of Information Engineering, Chinese Academy of Sciences \\
  \normalfont $^{2}$School of Cyber Security, University of Chinese Academy of Sciences \\
  \normalfont $^{3}$MiLM Plus, Xiaomi Inc. \\
  \normalfont $^{4}$Department of Computer Science, Brown University
}
\hypersetup{pdftitle={OmniStepBench: Deployment-Aware Agent Evaluation across Modalities and Task Complexity},
  pdfauthor={Yu Liu; Zhilin Liu; Zhiwei Yang; Shaojie Zhang; Zheyuan Deng; Tingwei Huang; Zhenbo Luo; Lei Jiang; Yanbing Liu; Pei Fu},
  pdfsubject={arXiv preprint}}

\begin{document}
\maketitle
\begingroup
\renewcommand{\thefootnote}{\fnsymbol{footnote}}
\footnotetext[1]{These authors contributed equally.}
\footnotetext[2]{Corresponding author.}
\endgroup
\begin{abstract}
    As large language models evolve from question-answering systems into general-purpose agents, evaluation must move beyond static answer correctness to assess multimodal perception, multi-step execution, tool use, and artifact delivery. However, existing benchmarks are often tied to specific task types, execution environments, or scoring protocols, limiting their comparability, interpretability, and reliability for deployment decisions. We introduce \textbf{OmniStepBench}, a benchmark for deployment-aware agent evaluation across modalities and task complexity. Built on a shared OpenClaw execution environment, OmniStepBench organizes 233 tasks selected and adapted from 22 source benchmarks into a $2\times3$ workload matrix defined by input modality and execution form, and evaluates them under a unified contract-based protocol with evidence-based score auditing. We evaluate 30 commercial API models and 12 locally deployed open-weight models (42 models in total) over 9{,}218 applicable model--task runs, reporting accuracy and token consumption alongside reference costs for API models. Results show workload-dependent rankings, distinct accuracy--cost trade-offs between text and multimodal tasks, and competitive but weaker overall performance from local models. These findings support deployment decisions based on workload profiles, deployment mode, and accuracy--cost trade-offs rather than a single aggregate score.
\end{abstract}

\section{Introduction}
\label{sec:intro}

\begin{figure*}[!t]
\centering
\includegraphics[width=1\linewidth]{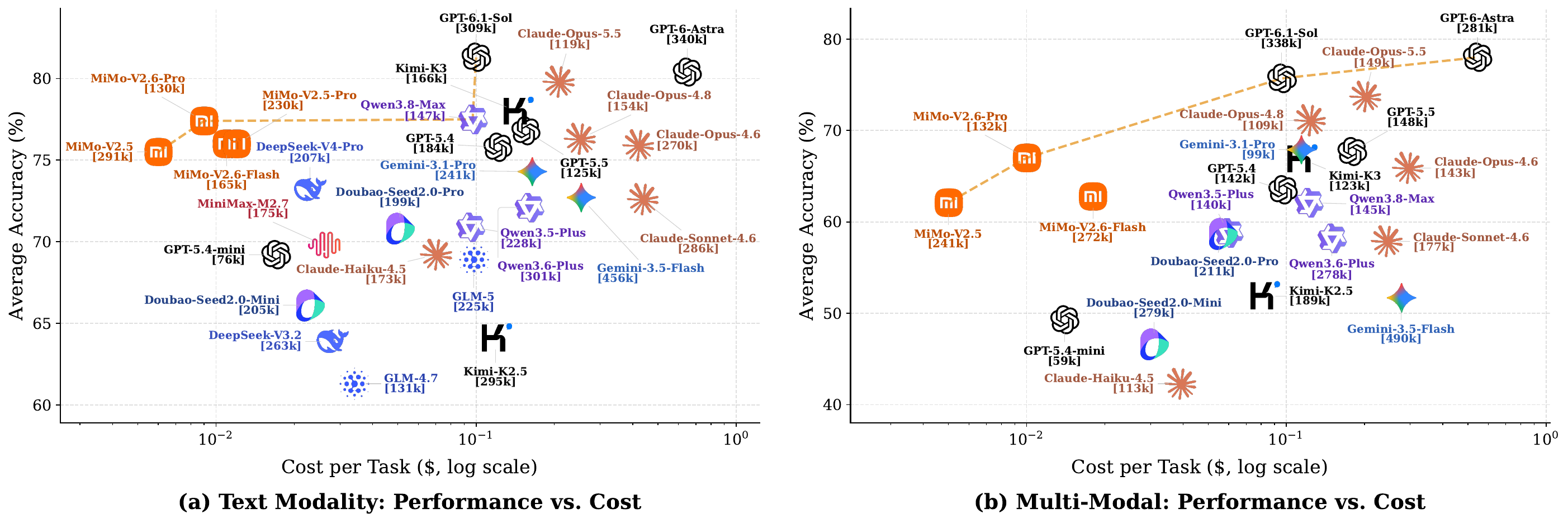}
\caption{Accuracy--cost trade-offs for API models on (a) text and (b) multimodal tasks.
Bracketed labels indicate average token consumption per task (in thousands); cost is reported in USD per task on a logarithmic scale.
Dashed lines connect the Pareto-optimal models among those shown:
MiMo-V2.5, MiMo-V2.6-Pro, Qwen3.8-Max, and GPT-6.1-Sol in (a);
and MiMo-V2.5, MiMo-V2.6-Pro, GPT-6.1-Sol, and GPT-6-Astra in (b).}
\label{fig:acc-cost}
\end{figure*}

Frontier large language models (LLMs) are evolving from standalone question-answering systems into core decision-making components of general-purpose agent systems. These agents must perceive, reason, plan, use tools, and deliver artifacts in dynamic workspaces, rather than merely generate an answer string~\citep{wei2022chain,schick2023toolformer}. Accordingly, model evaluation is shifting from static input--output performance to end-to-end execution under constraints on tools, context, time, and cost.

However, existing evaluation frameworks do not adequately support this transition. Traditional knowledge, reasoning, and long-context benchmarks primarily assess answer correctness under static inputs~\citep{hendrycks2021measuring,rein2023gpqa,bai2024longbench2}. Multimodal benchmarks incorporate visual inputs but still largely follow a static input--output paradigm~\citep{yue2024mmmu,wang2024charxiv}. Tool-use and end-to-end agent benchmarks, by contrast, are often confined to specific command-line, web, or desktop environments~\citep{terminalbench2026,xie2024osworld,mialon2023gaia,liu2023agentbench,zhou2023webarena,yao2024taubench}. Moreover, these benchmarks employ different runtimes, tool interfaces, task protocols, and scoring criteria, making their results difficult to compare directly. Recent studies have improved evaluation observability through execution trajectories, environment snapshots, and audit logs~\citep{claweval2026,clawevallive2026}. Nevertheless, these studies still primarily report overall success rates within specific environments. They therefore provide limited insight into model behavior across modalities, execution processes, and degrees of tool dependence. As a result, they offer limited support for comparable, interpretable, and reliable model selection in practice.

We argue that agent evaluation can inform deployment decisions only by addressing three interrelated questions, rather than by simply increasing benchmark scale. First, \textbf{are the results interpretable?} Agent tasks differ substantially in input modality, execution process, and tool dependence. A single aggregate score can obscure model strengths and weaknesses across workloads. Second, \textbf{are comparisons attributable?} When tasks rely on different agent frameworks, tool interfaces, and execution protocols, observed differences conflate model capability with runtime compatibility. Third, \textbf{are the scores trustworthy?} Open-ended agent tasks often involve multi-step operations and artifact delivery. Matching only the final text cannot verify whether a task was actually completed. Moreover, an LLM judge may incorrectly award positive credit when tools were not executed, the task timed out, or required artifacts were missing. Together, these issues call for a deployment-oriented benchmark that unifies \textbf{workload organization, execution environment control, and outcome verification}.

To address these requirements, we propose OmniStepBench, a benchmark for deployment-aware agent evaluation across modalities and task complexity. We select and adapt 233 tasks from 22 existing benchmarks through feasibility, difficulty, and representativeness filters, then organize them into a $2\times3$ matrix spanning text or multimodal input and single-step, multi-step, or tool-mediated execution. A shared OpenClaw runtime~\citep{openclaw2026} standardizes the agent loop, tools, workspace, and logs, while task contracts specify required artifacts, scorers, and time budgets. Outcomes are evaluated automatically, by an LLM judge, or with a hybrid scorer; deterministic checks and a meta-judge audit unsupported positive scores. We evaluate 30 commercial API models and 12 locally deployed open-weight models over 9{,}218 applicable model--task runs, reporting accuracy, token usage, and reference API costs. Results show workload-dependent rankings, distinct text--multimodal accuracy--cost frontiers, and competitive but weaker overall performance from local models. The audit identifies and removes 143 positive scores unsupported by execution evidence.

Our main contributions are as follows:
\begin{itemize}
    \item
    We introduce \textbf{OmniStepBench}, comprising 233 tasks adapted from 22 source benchmarks and organized under a unified runtime into a $2\times3$ workload matrix defined by modality and execution form.

    \item
    We evaluate 42 models (30 commercial API and 12 locally deployed open-weight models) over 9{,}218 model--task runs, revealing workload-dependent rankings, distinct accuracy--cost trade-offs across text and multimodal workloads, and the strengths and limitations of local deployment.

    \item
    We characterize unsupported positive scores in agent evaluation and introduce an evidence-based audit that verifies suspicious scores against execution trajectories and final artifacts.
\end{itemize}

\section{Related Work}

\begin{figure*}[!t]
\centering
\includegraphics[width=\textwidth]{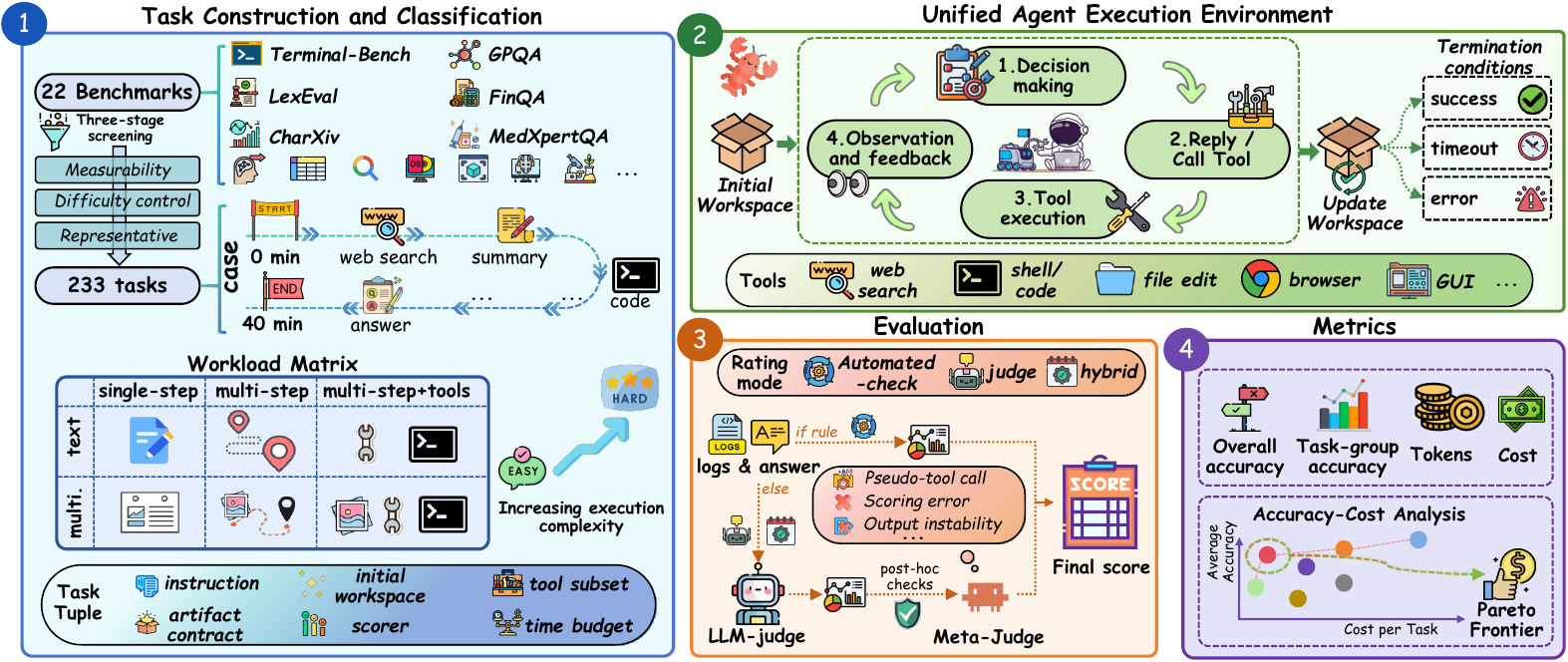}
\caption{
Overview of \benchname{}. The benchmark comprises four stages: 
(1) constructing 233 tasks from 22 source benchmarks and organizing them by modality and execution form;
(2) executing all models in a shared agent environment;
(3) applying automated, LLM-based, or hybrid scoring with evidence-based auditing; and
(4) reporting overall and workload-level accuracy, token usage, reference cost, and accuracy--cost frontiers.
}
\label{fig:overview}
\end{figure*}

\subsection{Agent Benchmarks and Capability Sources}

Agent evaluation is shifting from static question answering to execution in operational environments. Terminal-Bench~\citep{terminalbench2026}, OSWorld~\citep{xie2024osworld}, and MMSearch~\citep{jiang2024mmsearch} respectively assess command-line workflows, GUI interaction, and multimodal search, but remain tied to specific domains or environments.
OpenClaw has further motivated general-agent benchmarks. Claw-Eval~\citep{claweval2026} emphasizes trace-based trustworthy evaluation; ClawBench~\citep{clawbench2026} targets real-time web interaction; ClawArena~\citep{clawarena2026} studies conflict reasoning; WildClawBench~\citep{wildclawbench2026} and ZClawBench~\citep{zclawbench2026} introduce in-the-wild tool-use tasks; and ClawsBench~\citep{clawsbench2026} jointly evaluates capability and safety. However, differing runtimes, task taxonomies, and scoring protocols hinder systematic comparison under a unified framework.

\subsection{LLM-as-a-Judge in Agent Evaluation}

Following its validation on MT-Bench and Chatbot Arena~\citep{zheng2023judging}, LLM-as-a-Judge has been applied to code~\citep{jimenez2024swe,jiang2026codejudgebench}, multimodal~\citep{chen2024mllm}, dialogue~\citep{dubois2024length}, and security evaluation~\citep{shao2026towards}. Agent tasks are harder to judge because they involve long-horizon tool use, dynamic interactions, and artifact verification~\citep{li2025generation}.
Prior work shows that judges may infer success from plausible outputs without verifying execution evidence, causing ``hallucinated correctness''~\citep{you2026agent,gurram2026evaluating}. Claw-Eval~\citep{claweval2026} likewise reports substantial misses without audit logs. Yet systematic characterization and mitigation of such hallucinations remain limited.

\section{OmniStepBench}

    As illustrated in Figure~\ref{fig:overview}, \benchname{} consists of four components: workload construction, shared agent execution, reliable scoring, and deployment-oriented metrics. It organizes 233 tasks from 22 source benchmarks into a $2\times3$ workload matrix and evaluates all models under a unified execution protocol. Performance is assessed using automated, LLM-based, or hybrid scoring with evidence-based auditing, and reported alongside token usage and reference cost.

\subsection{Problem Definition}
    
    Unlike conventional benchmarks that evaluate only final responses, agentic tasks require models to interact with an execution environment and produce verifiable outputs. We formalize each task as
    \begin{equation}
    \tau_i =
    \bigl(
    p_i,\ \mathcal{W}_i^{0},\ \mathcal{T}_i,\ \mathcal{C}_i,\ G_i,\ t_i^{\max}
    \bigr),
    \end{equation}
    where $p_i$ is the task instruction, $\mathcal{W}_i^{0}$ the initial workspace, $\mathcal{T}_i$ the available tool set, $\mathcal{C}_i$ the artifact contract, $G_i$ the task-specific scorer, and $t_i^{\max}$ the execution-time budget.
    The artifact contract specifies the required output paths and their validity predicates:
    \begin{equation}
    \begin{aligned}
    \mathcal{C}_i &= \bigl\{(a_{i,j},\phi_{i,j})\bigr\}_{j=1}^{|\mathcal{C}_i|},\\
    \phi_{i,j}&: \text{file bytes} \rightarrow \{0,1\}.
    \end{aligned}
    \end{equation}
    Let $\mathcal{W}^{*}[a]$ denote the content at path $a$ in the final workspace. Complete contract satisfaction is defined as
    \begin{equation}
    \operatorname{Sat}(\mathcal{W}^{*},\mathcal{C}_i)
    =
    \prod_{j=1}^{|\mathcal{C}_i|}
    \mathbf{1}\!\left[\exists\,\mathcal{W}^{*}[a_{i,j}]\right]
    \phi_{i,j}\!\left(\mathcal{W}^{*}[a_{i,j}]\right).
    \end{equation}
    Under this formulation, a model operates within a shared execution environment to produce the required artifacts, which are evaluated by $G_i$. Task construction and scoring are described in
    \S\ref{sec:task-construction} and \S\ref{sec:evaluation-protocol}, respectively.
    
\subsection{Task Construction}
\label{sec:task-construction}

\subsubsection{Query Selection.}
    
    To construct a diverse and deployment-relevant evaluation suite, we draw candidate tasks from 22 publicly available benchmarks spanning reasoning, long-context understanding, question answering, instruction following, domain-specific analysis, code and shell workflows, desktop control, and multimodal understanding and search.
    
    We apply a three-stage curation procedure. First, the \textit{feasibility filter} removes tasks for which an explicit artifact contract $\mathcal{C}_i$ or a reliable scorer $G_i$ cannot be defined. Second, the \textit{difficulty filter} excludes overly simple tasks. For single-step tasks, we run Qwen3.5-27B three times and remove tasks solved correctly in all three trials. Where source difficulty annotations are available, we retain only tasks labeled \textit{medium} or \textit{hard}. Finally, the \textit{representativeness filter} manually selects tasks according to the category structure of each source benchmark, preserving broad coverage while avoiding over-representation of any single category. This procedure yields 233 tasks from the 22 source benchmarks.
    
\subsubsection{Workload Taxonomy.}
    
    Rather than organizing results by source benchmark, we group tasks along two workload dimensions: input modality and execution form. The modality dimension $\mathcal{M}$ distinguishes text-only and multimodal inputs, while the execution-form dimension $\mathcal{K}$ distinguishes single-step, multi-step, and multi-step-plus-tools tasks. Their Cartesian product defines six workload groups:
    
    \begin{align}
    \mathcal{G} &= \mathcal{M} \times \mathcal{K}, \\
    \mathcal{M} &= \{\text{text},\ \text{multimodal}\}, \\
    \mathcal{K} &= \{\text{single-step},\ \text{multi-step},\notag\\
    &\qquad\text{multi-step+tools}\}.
    \end{align}

    Operationally, we define a deterministic mapping
    $\varphi:\mathcal{B}\rightarrow\mathcal{G}$ that assigns each source benchmark to a single workload group according to its dominant modality and required execution form. A task $i$ originating from benchmark $b(i)\in\mathcal{B}$ inherits the corresponding group label:
    \begin{equation}
    g_i = \varphi\bigl(b(i)\bigr) = (m_i,k_i)\in\mathcal{G},
    \end{equation}
    where $m_i$ denotes the input modality and $k_i$ the required execution form. Let $T_g=\{i:g_i=g\}$ denote the set of tasks assigned to group $g$. These sets partition the benchmark, satisfying
    \begin{equation}
    \sum_{g\in\mathcal{G}} |T_g| = N,
    \end{equation}
    where $N=233$. The benchmark-to-group assignments and task counts are shown in the Supplementary Material.
    
    This taxonomy stratifies evaluation results along workload dimensions that are obscured by aggregate leaderboards. For example, a model may perform well on single-step text tasks while struggling with multi-step tool-mediated execution. Reporting performance across $\mathcal{G}$ therefore reveals workload-specific variation that would otherwise be hidden by a single aggregate score.
    
\subsection{Evaluation Protocol}
\label{sec:evaluation-protocol}

\subsubsection{Task Scoring.}

    Because \benchname{} contains heterogeneous task types, each task $\tau_i$ is assigned one of three scoring modes during benchmark construction:
    \[
    \eta_i \in \{\mathrm{auto},\mathrm{judge},\mathrm{hybrid}\},
    \]
    corresponding to automated, LLM-based, and hybrid scoring.
    
    \textbullet\ \textbf{Automated Scoring.}
    Automated scoring is used for tasks with deterministic answers or artifacts that can be verified through rule-based procedures. For single-step tasks, we check both the presence of required artifacts and the consistency of their contents with the ground truth. For multi-step tasks, we define task-specific scoring criteria over independently verifiable intermediate states and final outputs, weighted by their operational importance. This produces an automated score $s_{m,i}^{\mathrm{auto}}$ for model $m$ on task $i$.
    
    \textbullet\ \textbf{LLM-Based Scoring.}
    For tasks involving open-ended outputs or multiple semantic constraints that cannot be reliably evaluated using deterministic rules, we employ an LLM-as-a-Judge. A fixed judge model receives the task prompt $p_i$, a summary of the execution trajectory, the final artifacts in $\mathcal{W}_i^{*}$, and a manually specified task-level rubric. It returns criterion-level scores in a constrained format, which are aggregated using predefined rubric weights to obtain $s_{m,i}^{\mathrm{judge}}$.
    
    \textbullet\ \textbf{Hybrid Scoring.}
    Tasks containing both rule-verifiable and semantically judged criteria combine the two scoring modes. The weights
    $w_i^{\mathrm{auto}},w_i^{\mathrm{judge}}\in[0,1]$
    are fixed during task construction and satisfy
    $w_i^{\mathrm{auto}}+w_i^{\mathrm{judge}}=1$.
    The resulting task score is
    \begin{equation}
    s_{m,i} =
    \begin{cases}
    s_{m,i}^{\mathrm{auto}},
    & \eta_i=\mathrm{auto}, \\[4pt]
    s_{m,i}^{\mathrm{judge}},
    & \eta_i=\mathrm{judge}, \\[4pt]
    \begin{aligned}
    &w_i^{\mathrm{auto}}s_{m,i}^{\mathrm{auto}}\\
    &\quad + w_i^{\mathrm{judge}}s_{m,i}^{\mathrm{judge}},
    \end{aligned}
    & \eta_i=\mathrm{hybrid}.
    \end{cases}
    \end{equation}
    
    All task scores are normalized to $[0,1]$.

\subsubsection{Evidence-Based Audit.}

    Our pilot experiments revealed cases in which the primary judge awarded positive credit that was not supported by the execution trajectory or final artifacts. We refer to these cases as \emph{unsupported positive judgments}. Typical examples include nonzero scores assigned to empty outputs, executions without substantive operations, or runs missing required artifacts.
    
    To identify and correct such cases, we apply deterministic post-hoc rules to all scores involving the primary judge. Cases satisfying one or more trigger conditions are routed to a separate meta-judge from a different model family. The meta-judge receives the complete task prompt, execution trajectory, final artifacts, scoring rubric, and the primary judge's score and rationale. It then determines whether the awarded positive score is supported by the available execution evidence.
    
    If the meta-judge determines that the score is unsupported as a whole, the task score is set to zero; otherwise, the original score is retained. The audited score is defined as
    \begin{equation}
    s_{m,i}^{\mathrm{audit}}
    =
    (1-h_{m,i})s_{m,i},
    \end{equation}
    where $h_{m,i}\in\{0,1\}$ indicates whether the primary score is confirmed to be unsupported. Unflagged cases and flagged cases upheld by the meta-judge are assigned $h_{m,i}=0$. Detailed trigger rules, meta-judge prompts, case studies, and human--meta-judge agreement results are provided in the appendix.

\subsubsection{Metrics and Efficiency Accounting.}
    
    Let $\mathcal{I}_m$ denote the set of tasks applicable to model $m$, and let
    $\mathcal{I}_{m,g}=\mathcal{I}_m\cap T_g$
    denote its applicable tasks in workload group $g$. We report overall and group-level audited accuracy as
    \begin{align}
    \mathrm{Acc}(m)
    &=
    \frac{1}{|\mathcal{I}_m|}
    \sum_{i\in\mathcal{I}_m}
    s_{m,i}^{\mathrm{audit}}, \\
    \mathrm{Acc}_g(m)
    &=
    \frac{1}{|\mathcal{I}_{m,g}|}
    \sum_{i\in\mathcal{I}_{m,g}}
    s_{m,i}^{\mathrm{audit}}.
    \end{align}
    We report average token usage and estimated dollar cost per task over $\mathcal{I}_m$. Writing $u_{m,i}^{k}$ for the token count of kind $k\in\{\mathrm{in},\mathrm{out},\mathrm{cr},\mathrm{cw}\}$ (uncached input, output, cache-read, and cache-write), the two efficiency metrics are
    \begin{equation}
    \mathrm{Token}(m)=\frac{1}{|\mathcal{I}_m|}\sum_{i\in\mathcal{I}_m}\sum_{k}u_{m,i}^{k},
    \end{equation}
    \begin{equation}
    \mathrm{Cost}(m)=\frac{1}{|\mathcal{I}_m|}\sum_{i\in\mathcal{I}_m}\frac{1}{10^{6}}\sum_{k}\alpha_m^{k}u_{m,i}^{k},
    \end{equation}
    where $\alpha_m^{k}$ is the per-million-token price of kind $k$, taken from model-specific reference rates. When a provider exposes no separate cache-read or cache-write rate, those tokens are billed at the input price ($\alpha_m^{\mathrm{cr}}$ or $\alpha_m^{\mathrm{cw}}=\alpha_m^{\mathrm{in}}$). For locally deployed models, any hosted-inference price is only a third-party reference rather than the measured cost of local vLLM serving; accuracy--cost frontier analyses are therefore based on API reference costs. Full per-model rates are listed in the supplementary material.

\definecolor{MacRed}{HTML}{FF8A8A}   
\definecolor{MacBlue}{HTML}{8AB8FF}  

\begin{table*}[!t]
\centering
\small
\setlength{\tabcolsep}{6pt}
\resizebox{\textwidth}{!}{
\begin{tabular}{l ccc >{\bfseries}c ccc >{\bfseries}c}
\toprule
& \multicolumn{4}{c}{\textbf{Text}} & \multicolumn{4}{c}{\textbf{Multi-Modal}} \\
\cmidrule(lr){2-5}\cmidrule(lr){6-9}
\textbf{Model} & \textbf{SS} & \textbf{MS} & \textbf{MT} & \textbf{Avg} & \textbf{SS} & \textbf{MS} & \textbf{MT} & \textbf{Avg} \\
\midrule
\rowcolor{gray!15}
\multicolumn{9}{l}{\textbf{\textit{API Deployment}}} \\
GPT-6-Astra~\citep{openai2026gpt6astra} & 96.0 & 82.8 & \underline{70.6} & \cellcolor{MacRed!93}\underline{80.3} & 70.0 & \textbf{94.2} & \textbf{67.6} & \cellcolor{MacBlue!96}\textbf{78.0} \\

GPT-6.1-Sol~\citep{openai2026gpt61sol} & 96.1 & \underline{84.0} & \textbf{71.3} & \cellcolor{MacRed!97}\textbf{81.2} & 70.0 & \underline{92.3} & \underline{62.9} & \cellcolor{MacBlue!90}\underline{75.7} \\

Claude-Opus-5.5~\citep{anthropic2026opus55} & \textbf{99.2} & \textbf{84.4} & 65.0 & \cellcolor{MacRed!91}\textit{79.6} & 75.0 & \underline{92.3} & 53.2 & \cellcolor{MacBlue!86}\textit{73.7} \\

Claude-Opus-4.8~\citep{anthropic2026opus48} & \underline{98.8} & 78.1 & \textit{65.3} & \cellcolor{MacRed!75}76.3 & \underline{85.0} & 69.2 & \textit{61.8} & \cellcolor{MacBlue!80}71.1 \\

Kimi-K3~\citep{moonshotai2026kimik3} & 95.3 & 82.9 & 64.0 & \cellcolor{MacRed!83}78.0 & 80.0 & 69.2 & 54.1 & \cellcolor{MacBlue!68}66.9 \\

MiMo-V2.6-Pro~\citep{xiaomi2026mimov26pro} & 91.5 & 82.9 & 63.6 & \cellcolor{MacRed!81}77.4 & \underline{85.0} & 69.2 & 50.4 & \cellcolor{MacBlue!68}67.0 \\

GPT-5.5~\citep{openai2026gpt55} & 94.1 & \textit{83.6} & 59.8 & \cellcolor{MacRed!78}76.8 & \textbf{90.0} & 75.0 & 42.4 & \cellcolor{MacBlue!70}67.7 \\

Claude-Opus-4.6~\citep{anthropic2026opus46} & 91.5 & 83.4 & 58.3 & \cellcolor{MacRed!73}75.9 & 75.0 & 76.2 & 48.1 & \cellcolor{MacBlue!65}65.9 \\

Qwen3.8-Max~\citep{qwen2026qwen38max} & 89.7 & 82.6 & 65.0 & \cellcolor{MacRed!81}77.5 & 80.0 & 61.5 & 48.4 & \cellcolor{MacBlue!55}62.1 \\

Gemini-3.1-Pro~\citep{geminiteam2026gemini31pro} & \textit{96.4} & 80.7 & 56.1 & \cellcolor{MacRed!65}74.3 & \underline{85.0} & 88.5 & 32.7 & \cellcolor{MacBlue!71}67.9 \\

GPT-5.4~\citep{openai2026gpt54} & 92.1 & 82.4 & 59.6 & \cellcolor{MacRed!72}75.8 & 82.5 & 71.2 & 40.3 & \cellcolor{MacBlue!58}63.5 \\

MiMo-V2.6-Flash~\citep{xiaomi2026mimov26flash} & 84.2 & 81.8 & 63.8 & \cellcolor{MacRed!74}76.0 & 80.0 & 65.4 & 46.5 & \cellcolor{MacBlue!57}62.8 \\

MiMo-V2.5~\citep{xiaomi2026mimov25} & 94.7 & 82.5 & 57.5 & \cellcolor{MacRed!71}75.5 & 80.0 & 66.7 & 43.0 & \cellcolor{MacBlue!55}62.1 \\

Claude-Sonnet-4.6~\citep{anthropic2026sonnet46} & 89.2 & 76.8 & 59.9 & \cellcolor{MacRed!58}72.6 & 75.0 & 59.6 & 42.3 & \cellcolor{MacBlue!45}57.9 \\

Qwen3.6-Plus~\citep{qwen2026qwen36plus} & 95.5 & 77.7 & 54.8 & \cellcolor{MacRed!55}72.1 & 75.0 & 61.5 & 41.2 & \cellcolor{MacBlue!46}58.2 \\

Qwen3.5-Plus~\citep{qwen2026qwen35} & 89.5 & 77.3 & 53.9 & \cellcolor{MacRed!50}70.9 & 70.0 & 69.2 & 39.1 & \cellcolor{MacBlue!47}58.8 \\

Doubao-Seed2.0-Pro~\citep{bytedanceseed2026seed20} & 95.9 & 77.4 & 51.1 & \cellcolor{MacRed!49}70.8 & 75.0 & 61.5 & 42.6 & \cellcolor{MacBlue!47}58.7 \\

Gemini-3.5-Flash~\citep{googledeepmind2026gemini35} & 94.9 & 80.9 & 51.7 & \cellcolor{MacRed!58}72.7 & \underline{85.0} & 50.0 & 26.7 & \cellcolor{MacBlue!30}51.7 \\

GPT-5.4-mini~\citep{openai2026gpt54mini} & 77.8 & 79.1 & 50.3 & \cellcolor{MacRed!42}69.2 & 70.0 & 46.2 & 35.9 & \cellcolor{MacBlue!25}49.3 \\

Claude-Haiku-4.5~\citep{anthropic2025haiku45} & 89.1 & 76.7 & 50.1 & \cellcolor{MacRed!42}69.2 & 50.5 & 46.2 & 31.6 & \cellcolor{MacBlue!15}42.3 \\

MiMo-V2-Omni~\citep{xiaomi2026mimov2omni} & 70.4 & 75.4 & 51.5 & \cellcolor{MacRed!32}66.7 & 75.0 & 34.6 & 34.7 & \cellcolor{MacBlue!20}46.0 \\

Kimi-K2.5~\citep{moonshotai2026kimik25} & 89.2 & 70.0 & 45.6 & \cellcolor{MacRed!22}64.1 & 65.0 & 54.4 & 38.7 & \cellcolor{MacBlue!31}51.9 \\

Doubao-Seed2.0-Mini~\citep{bytedanceseed2026seed20} & 75.7 & 78.9 & 42.5 & \cellcolor{MacRed!30}66.1 & 55.0 & 51.9 & 34.3 & \cellcolor{MacBlue!21}46.6 \\

MiMo-V2.5-Pro~\citep{xiaomi2026mimov25pro} & 91.6 & 81.7 & 61.2 & \cellcolor{MacRed!74}76.0 & --- & --- & --- & --- \\

DeepSeek-V4-Pro~\citep{deepseek2026v4} & 91.6 & 80.9 & 54.3 & \cellcolor{MacRed!60}73.2 & --- & --- & --- & --- \\

MiMo-V2-Pro~\citep{xiaomi2026mimov2pro} & 91.7 & 75.4 & 58.4 & \cellcolor{MacRed!52}71.6 & --- & --- & --- & --- \\

MiniMax-M2.7~\citep{minimax2026m27} & 80.9 & 77.8 & 53.0 & \cellcolor{MacRed!45}69.8 & --- & --- & --- & --- \\

GLM-5~\citep{zai2026glm5} & 84.9 & 75.0 & 53.3 & \cellcolor{MacRed!41}68.9 & --- & --- & --- & --- \\

DeepSeek-V3.2~\citep{deepseek2025v32} & 80.0 & 71.1 & 46.8 & \cellcolor{MacRed!21}63.9 & --- & --- & --- & --- \\

GLM-4.7~\citep{zai2025glm47} & 73.1 & 76.2 & 33.6 & \cellcolor{MacRed!12}61.3 & --- & --- & --- & --- \\

\midrule
\rowcolor{gray!15}
\multicolumn{9}{l}{\textbf{\textit{Local Deployment}}} \\

Qwen3.6-27B~\citep{qwen2026qwen3627b} & \underline{84.9} & \textbf{72.1} & \underline{37.2} & \cellcolor{MacRed!42}\textbf{61.8} & \textbf{75.0} & \underline{55.8} & \textbf{38.5} & \cellcolor{MacBlue!40}\textbf{55.1} \\

Qwen3.6-35B-A3B~\citep{qwen2026qwen3635b} & 79.3 & \underline{70.2} & 33.9 & \cellcolor{MacRed!32}\underline{59.0} & \underline{65.0} & \textbf{57.7} & \underline{32.7} & \cellcolor{MacBlue!30}\underline{51.0} \\

Qwen3.5-27B~\citep{qwen2026qwen35} & \textbf{85.8} & 59.9 & 27.0 & \cellcolor{MacRed!15}51.9 & 55.0 & 53.8 & \textit{24.5} & \cellcolor{MacBlue!18}\textit{43.9} \\

Qwen3.5-35B-A3B~\citep{qwen2026qwen35} & 66.2 & 57.9 & \textit{34.0} & \cellcolor{MacRed!12}50.8 & 60.0 & 38.5 & 22.9 & \cellcolor{MacBlue!10}39.0 \\

Qwen3-VL-8B-Instruct~\citep{bai2025qwen3vltechnicalreport} & 35.7 & 58.2 & 28.9 & \cellcolor{MacRed!5}45.5 & 40.0 & 25.0 & 22.3 & \cellcolor{MacBlue!5}28.3 \\

Qwen2.5-VL-32B-Instruct~\citep{bai2025qwen2} & 53.3 & \textit{61.8} & 20.1 & \cellcolor{MacRed!7}46.6 & 15.0 & 34.6 & 13.0 & \cellcolor{MacBlue!2}21.5 \\

Qwen3-VL-32B-Instruct~\citep{bai2025qwen3vltechnicalreport} & 48.3 & 48.7 & 25.2 & \cellcolor{MacRed!2}40.6 & 55.0 & 34.6 & 17.4 & \cellcolor{MacBlue!7}34.3 \\

Gemma-4-26B-A4B-it~\citep{gemmateam2026gemma4} & \textit{83.0} & 40.6 & 19.3 & \cellcolor{MacRed!2}38.6 & 14.0 & \underline{55.8} & 20.7 & \cellcolor{MacBlue!6}31.6 \\

Qwen3.5-9B~\citep{qwen2026qwen35} & 56.4 & 32.9 & 27.1 & \cellcolor{MacRed!2}33.8 & \textit{62.5} & 28.8 & 16.2 & \cellcolor{MacBlue!7}33.9 \\

Qwen2.5-VL-7B-Instruct~\citep{bai2025qwen2} & 28.0 & 35.2 & 19.7 & \cellcolor{MacRed!2}29.0 & 10.0 & 7.7 & 4.0 & \cellcolor{MacBlue!2}7.0 \\

Gemma-3-27B-it~\citep{kamath2025gemma3} & 42.4 & 29.5 & 14.5 & \cellcolor{MacRed!2}26.0 & 25.0 & 13.5 & 0.0 & \cellcolor{MacBlue!2}12.0 \\

Qwen3-Coder-30B-A3B-Instruct~\citep{yang2025qwen3} & 73.9 & 56.3 & \textbf{37.7} & \cellcolor{MacRed!16}\textit{52.1} & --- & --- & --- & --- \\

\bottomrule
\end{tabular}}

\caption{Per-model accuracy (\%) across the six \benchname{} task groups. Averages are task-count-weighted (Text: 20/87/55; Multi-Modal: 20/26/25). SS, MS, and MT denote Single-Step, Multi-Step, and Multi-Step+Tools. \textbf{Bold}, \underline{underlined}, and \textit{italicized} values denote first, second, and third place within each deployment mode (competition ranking); ``---'' marks text-only models not evaluated on multimodal tasks.}
\label{tab:acc}
\end{table*}
\definecolor{MacRed}{HTML}{FF8A8A}   
\definecolor{MacBlue}{HTML}{8AB8FF}  

\begin{table}[t]
\centering
\footnotesize
\setlength{\tabcolsep}{3pt}
\resizebox{\linewidth}{!}{
\begin{adjustbox}{max width=\columnwidth}
\begin{tabular}{@{}lcccc@{}}
\toprule
 & \multicolumn{2}{c}{\textbf{Tokens/task}} & \multicolumn{2}{c}{\textbf{Cost (USD)/task}} \\
\cmidrule(lr){2-3}\cmidrule(lr){4-5}
Model & \textbf{Text Avg} & \textbf{MM Avg} & \textbf{Text Avg} & \textbf{MM Avg} \\
\midrule
GPT-6-Astra & \cellcolor{MacRed!60}345k & \cellcolor{MacBlue!51}281k & \cellcolor{MacRed!100}0.654 & \cellcolor{MacBlue!100}0.544 \\
GPT-6.1-Sol & \cellcolor{MacRed!55}313k & \cellcolor{MacBlue!60}338k & \cellcolor{MacRed!22}0.100 & \cellcolor{MacBlue!29}0.096 \\
Claude-Opus-5.5 & \cellcolor{MacRed!22}\underline{119k} & \cellcolor{MacBlue!30}149k & \cellcolor{MacRed!41}0.207 & \cellcolor{MacBlue!56}0.201 \\
Claude-Opus-4.8 & \cellcolor{MacRed!32}154k & \cellcolor{MacBlue!26}\textit{109k} & \cellcolor{MacRed!60}0.250 & \cellcolor{MacBlue!40}0.123 \\
Kimi-K3 & \cellcolor{MacRed!35}166k & \cellcolor{MacBlue!29}123k & \cellcolor{MacRed!40}0.144 & \cellcolor{MacBlue!38}0.114 \\
MiMo-V2.6-Pro & \cellcolor{MacRed!27}130k & \cellcolor{MacBlue!31}132k & \cellcolor{MacRed!7}\textit{0.009} & \cellcolor{MacBlue!8}\textit{0.010} \\
GPT-5.5 & \cellcolor{MacRed!25}\textit{125k} & \cellcolor{MacBlue!35}148k & \cellcolor{MacRed!42}0.155 & \cellcolor{MacBlue!52}0.179 \\
Claude-Opus-4.6 & \cellcolor{MacRed!56}270k & \cellcolor{MacBlue!34}143k & \cellcolor{MacRed!80}0.419 & \cellcolor{MacBlue!80}0.293 \\
Qwen3.8-Max & \cellcolor{MacRed!31}147k & \cellcolor{MacBlue!34}145k & \cellcolor{MacRed!30}0.097 & \cellcolor{MacBlue!40}0.122 \\
Gemini-3.1-Pro & \cellcolor{MacRed!51}241k & \cellcolor{MacBlue!24}\underline{99k} & \cellcolor{MacRed!44}0.164 & \cellcolor{MacBlue!38}0.114 \\
GPT-5.4 & \cellcolor{MacRed!38}184k & \cellcolor{MacBlue!34}142k & \cellcolor{MacRed!35}0.121 & \cellcolor{MacBlue!34}0.097 \\
MiMo-V2.6-Flash & \cellcolor{MacRed!34}165k & \cellcolor{MacBlue!61}272k & \cellcolor{MacRed!7}0.011 & \cellcolor{MacBlue!11}0.018 \\
MiMo-V2.5 & \cellcolor{MacRed!61}291k & \cellcolor{MacBlue!57}241k & \cellcolor{MacRed!5}\textbf{0.006} & \cellcolor{MacBlue!6}\underline{0.005} \\
Claude-Sonnet-4.6 & \cellcolor{MacRed!60}286k & \cellcolor{MacBlue!42}177k & \cellcolor{MacRed!83}0.436 & \cellcolor{MacBlue!68}0.243 \\
Qwen3.6-Plus & \cellcolor{MacRed!63}301k & \cellcolor{MacBlue!65}278k & \cellcolor{MacRed!43}0.160 & \cellcolor{MacBlue!46}0.150 \\
Qwen3.5-Plus & \cellcolor{MacRed!48}228k & \cellcolor{MacBlue!33}140k & \cellcolor{MacRed!30}0.095 & \cellcolor{MacBlue!24}0.060 \\
Doubao-Seed2.0-Pro & \cellcolor{MacRed!42}199k & \cellcolor{MacBlue!50}211k & \cellcolor{MacRed!20}0.051 & \cellcolor{MacBlue!23}0.057 \\
Gemini-3.5-Flash & \cellcolor{MacRed!80}456k & \cellcolor{MacBlue!85}490k & \cellcolor{MacRed!61}0.253 & \cellcolor{MacBlue!76}0.277 \\
GPT-5.4-mini & \cellcolor{MacRed!15}\textbf{76k} & \cellcolor{MacBlue!15}\textbf{59k} & \cellcolor{MacRed!10}0.017 & \cellcolor{MacBlue!10}0.014 \\
Claude-Haiku-4.5 & \cellcolor{MacRed!36}173k & \cellcolor{MacBlue!27}113k & \cellcolor{MacRed!25}0.070 & \cellcolor{MacBlue!18}0.039 \\
MiMo-V2-Omni & \cellcolor{MacRed!55}264k & \cellcolor{MacBlue!32}134k & \cellcolor{MacRed!6}\underline{0.008} & \cellcolor{MacBlue!5}\textbf{0.003} \\
Kimi-K2.5 & \cellcolor{MacRed!62}295k & \cellcolor{MacBlue!45}189k & \cellcolor{MacRed!35}0.119 & \cellcolor{MacBlue!30}0.082 \\
Doubao-Seed2.0-Mini & \cellcolor{MacRed!43}205k & \cellcolor{MacBlue!65}279k & \cellcolor{MacRed!12}0.023 & \cellcolor{MacBlue!15}0.031 \\
MiMo-V2.5-Pro & \cellcolor{MacRed!48}230k & --- & \cellcolor{MacRed!8}0.012 & --- \\
DeepSeek-V4-Pro & \cellcolor{MacRed!43}207k & --- & \cellcolor{MacRed!12}0.023 & --- \\
MiMo-V2-Pro & \cellcolor{MacRed!49}234k & --- & \cellcolor{MacRed!9}0.014 & --- \\
MiniMax-M2.7 & \cellcolor{MacRed!36}175k & --- & \cellcolor{MacRed!13}0.026 & --- \\
GLM-5 & \cellcolor{MacRed!47}225k & --- & \cellcolor{MacRed!31}0.098 & --- \\
DeepSeek-V3.2 & \cellcolor{MacRed!55}263k & --- & \cellcolor{MacRed!14}0.028 & --- \\
GLM-4.7 & \cellcolor{MacRed!27}131k & --- & \cellcolor{MacRed!16}0.034 & --- \\
\bottomrule
\end{tabular}
\end{adjustbox}}
\caption{Per-model token consumption and reference dollar cost per task on \benchname{}. Averages are task-count-weighted. Full SS, MS, and MT results are provided in Appendix~\ref{sec:api-efficiency-details}.}
\label{tab:cost}
\end{table}

\section{Experiments}
\label{sec:experiments}

\subsection{Experimental Setup}

\noindent \textbf{Models Evaluated.}
    We evaluate 42 models: 30 commercial API models from 10 families and 12 open-weight models served locally with vLLM on a single NVIDIA RTX PRO 6000 GPU (96 GB VRAM). Local models use a common FP16, 32K-context configuration---the largest stable setting shared by all models on one GPU---without quantization or multi-GPU serving. API--local comparisons therefore reflect end-to-end deployment performance rather than context-matched intrinsic capability. All runs use the same OpenClaw runtime and default inference settings, with isolated sessions, workspaces, artifacts, and logs.

\noindent \textbf{Infrastructure.}
    All models use the same OpenClaw tool suite, comprising the default tools, \texttt{web\_search}, \texttt{web\_fetch}, and nine Clawhub skills for reasoning, automation, and general utilities. Tool and skill visibility, permissions, and interfaces are identical across models; the complete configuration is provided in the supplementary material.

\noindent \textbf{Evaluation Metrics.}
    We use accuracy as the primary capability metric, reporting overall accuracy and accuracy for six workload groups: Text-SS, Text-MS, Text-MT, MM-SS, MM-MS, and MM-MT. Efficiency is measured by average token usage and estimated USD cost per task, computed from measured input/output tokens using model-specific reference rates; pricing details are provided in the supplementary material. For LLM-based scoring, we use Qwen3.5-VL-Plus as the primary judge and Claude-Opus-4.6 as the meta-judge.

\subsection{Main Results}
    Table~\ref{tab:acc} reports accuracy across the six \benchname{} task groups, with models grouped by deployment mode. Text-only models are evaluated only on text tasks; token usage and reference API cost are reported for all API models in Table~\ref{tab:cost}, with local-model details in the Supplementary. Figure~\ref{fig:acc-cost} visualizes the API accuracy--cost trade-off. Three patterns emerge. First, capability is strongly workload-dependent. Second, execution complexity creates a consistent pressure on accuracy. Third, deployment constraints affect both the attainable accuracy and the dominant failure modes.
    
    \textbullet\ \textbf{Performance is workload-dependent.} No model dominates all six groups. Models that are strong on text are not necessarily the strongest on visual inputs, and the ordering changes again when tasks require multiple actions or external tools. These shifts reflect an interaction between input modality and execution form: a model may handle static reasoning well yet struggle to maintain state, invoke tools reliably, or verify artifacts over a longer trajectory. The spread between models is therefore informative, because aggregate scores can conceal complementary strengths in perception, reasoning, tool use, and artifact production. Model selection should consequently be guided by the target workload profile and its operational requirements, rather than by a single leaderboard score.
    
    \textbullet\ \textbf{Execution complexity amplifies the deployment gap.} On average, accuracy is highest for single-step tasks and declines as workload moves to multi-step and tool-mediated execution, with the lowest accuracy in multimodal tool-use settings. The strongest local models remain competitive on simpler and intermediate workloads, but the gap to API models widens on long-horizon tasks that require sustained state tracking, reliable tool invocation, and artifact completion. Smaller local models are more vulnerable to instability in the full agent loop. These results indicate that deployment feasibility depends on the workload's execution demands, not only on the model's standalone capability.

\begin{table}[!t]
\centering
\small
\begin{tabular}{lccc}
\hline
\textbf{Pattern} & \textbf{Local} & \textbf{API} & \textbf{Total} \\
\hline
Deadlock loop & 86 & 0 & 86 \\
Rubric leakage & 12 & 14 & 26 \\
Pseudo-tool-call credulity & 20 & 0 & 20 \\
Output instability & 5 & 2 & 7 \\
Timeout-with-credit & 1 & 3 & 4 \\
\hline
\textbf{Total} & \textbf{124} & \textbf{19} & \textbf{143} \\
\hline
\end{tabular}
\caption{Audit type of LLM Judge Hallucination Patterns}
\vspace{-2mm}
\label{tab:hallucination_patterns}
\end{table}

    \textbullet\ \textbf{Token use and monetary cost are weakly coupled.} API models span a wide range of token demand, and provider pricing changes the resulting cost ordering substantially. Some models spend many tokens while remaining inexpensive under their reference rates, whereas others achieve shorter trajectories at a much higher price. Token usage therefore measures execution demand, while dollar cost reflects the interaction between that demand and provider pricing; both are needed to assess practical efficiency.
    
    \textbullet\ \textbf{Accuracy--cost frontiers differ by modality.} Figure~\ref{fig:acc-cost} shows a relatively flat text frontier and a steeper multimodal frontier. Text workloads approach an accuracy plateau at comparatively low reference cost, whereas multimodal workloads retain larger gains as cost increases. The cost of a model is therefore not informative without its workload context, and deployment choices should consider modality together with expected accuracy and token demand.

\subsection{LLM Judge Reliability}

    LLM-based scoring extends evaluation to open-ended outputs and artifacts, but it can also assign positive credit when the execution evidence does not support the judgment. We therefore audit positive scores that satisfy deterministic trigger rules using a separate meta-judge. The meta-judge reviews the task prompt, execution trajectory, final artifacts, scoring rubric, and primary-judge rationale; scores judged unsupported are reset to zero, whereas supported scores are retained.

    Table~\ref{tab:hallucination_patterns} summarizes the 143 unsupported positive judgments identified by this procedure. Five recurring patterns emerge: deadlock loops, rubric leakage, pseudo-tool-call credulity, output instability, and timeout-with-credit. Deadlock loops are the most frequent pattern, followed by rubric leakage and pseudo-tool-call credulity. Together, these cases show that judge errors often arise when plausible textual reasoning is mistaken for completed execution, rather than from disagreement about the task answer alone.

    The pattern distribution also varies with deployment mode. Local runs account for most audited cases, dominated by deadlock loops and pseudo-tool-call credulity; neither pattern occurs in the API runs. Under the fixed 32K single-GPU configuration, these failures are consistent with model-dependent instability in long-horizon execution, although the concentration in a subset of local models indicates that task length alone is insufficient as an explanation. API runs instead show more rubric leakage and timeout-with-credit, suggesting different reliability pressures in hosted invocation. These observations characterize the specified deployment profile and should not be extrapolated directly to other runtimes, context limits, or serving conditions.

\begin{figure}[!t]
\centering
\includegraphics[width=1\linewidth]{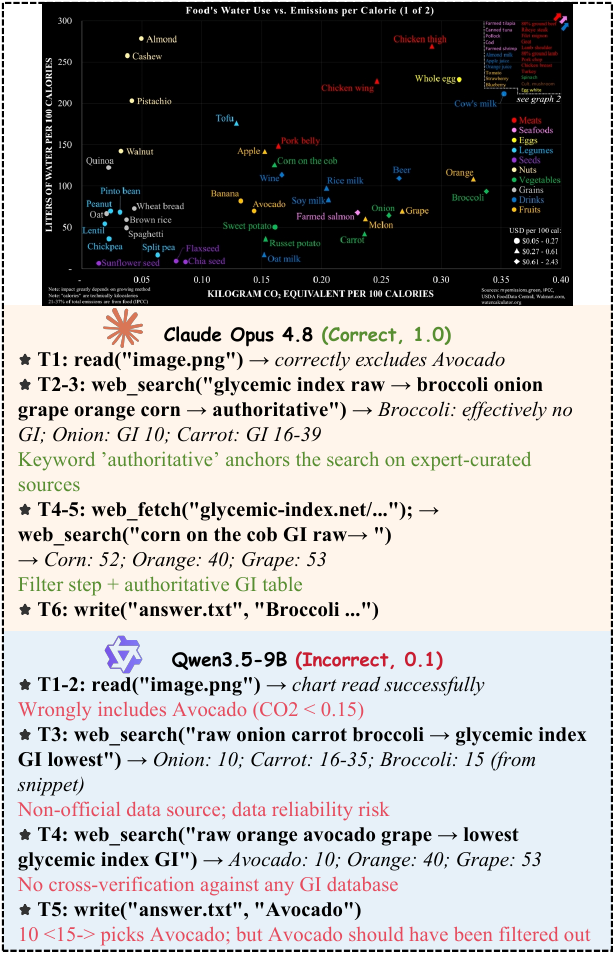}
\caption{
Multi-Step Visual Search on a Nutrition Scatter Chart.
\textbf{Example Task:} Among all Fruits and Vegetables with CO\textsubscript{2}~$>0.15$~kg per 100 calories and price tier of either Highest or Medium, which single food has the lowest Glycemic Index (GI)?
}
\label{case}
\end{figure}

\subsection{Case Study}
    Figure~\ref{case} illustrates a representative multimodal agent failure. The task requires filtering foods by two visual constraints before retrieving their glycemic-index values. Qwen3.5-9B misreads the chart and retains \emph{Avocado} despite the CO\textsubscript{2} threshold; this early error changes the candidate set and leads to the wrong answer. It also relies on unverified sources. In contrast, Claude Opus 4.8 applies both constraints and checks retrieved values against authoritative sources. With identical tools and runtime, the contrast highlights differences in visual grounding, constraint tracking, and evidence verification, as well as how early errors propagate through multi-step trajectories. Further representative cases appear in the Supplementary Material.
\section{Conclusion}

We present \benchname{}, a deployment-aware benchmark of 233 tasks organized by modality and execution complexity. Evaluations across 42 models show workload-dependent rankings, distinct text--multimodal accuracy--cost trade-offs, and a persistent gap between local and API deployment on demanding agent tasks. Evidence-based auditing further identifies unsupported positive judgments. Together, these results establish \benchname{} as a unified framework for workload-specific and reliable agent evaluation. Future work will extend the benchmark across multiple agent runtimes and broader real-world task distributions.

\FloatBarrier
\section*{Limitations}

Because evaluating all models across multiple agent frameworks would substantially increase API and compute costs, we conduct the study in a single execution environment, OpenClaw. The shared runtime, tool interface, workspace abstraction, and logging protocol make the within-benchmark comparison controlled, but the results may not transfer unchanged to frameworks with different tool semantics, orchestration strategies, or resource budgets. In particular, the comparison between API and locally served models reflects their specified deployment profiles rather than context-matched intrinsic capability.

The benchmark contains 233 tasks adapted from 22 source benchmarks and selected using feasibility, difficulty, and representativeness criteria. This curation provides structured coverage of six workload groups, but it cannot represent the full diversity of real-world agent tasks. Aggregate results may therefore depend on the task mix and group distribution, and should be interpreted as evidence for the evaluated workload profile rather than as a universal ranking of agent models.

\bibliography{references}
\clearpage
\appendix
\setcounter{topnumber}{4}
\setcounter{bottomnumber}{2}
\setcounter{totalnumber}{6}
\setcounter{dbltopnumber}{3}
\renewcommand{\topfraction}{0.95}
\renewcommand{\bottomfraction}{0.9}
\renewcommand{\textfraction}{0.05}
\renewcommand{\dbltopfraction}{0.95}
\renewcommand{\floatpagefraction}{0.75}
\renewcommand{\dblfloatpagefraction}{0.75}
\setlength{\textfloatsep}{12pt plus 2pt minus 2pt}
\setlength{\dbltextfloatsep}{12pt plus 2pt minus 2pt}
\setlength{\floatsep}{8pt plus 2pt minus 2pt}
\setlength{\dblfloatsep}{6pt plus 2pt minus 2pt}
\raggedbottom
\section{Details of Experimental Setup} \label{sec:exp-setup}

This section complements the experimental setup summarized in the main paper with the details required for reproduction.
We focus on inference configuration and the tool \& skill suite;
in addition, the models used and their per-token prices are listed in Table~\ref{tab:openrouter-prices}~(\S\ref{sec:openrouter-pricing}) and the judge/meta-judge prompts are reproduced verbatim in Figure~\ref{fig:judge-prompt} and Figure~\ref{fig:meta-judge-system}.

\subsection{Inference Configuration}
All models are evaluated under the OpenClaw~\citep{openclaw2026} defaults to avoid per-model tuning bias.
The random seed is fixed to 42 across all decode calls.
Local models are served with vLLM~\citep{kwon2023efficient} on the NVIDIA RTX PRO 6000 workstation~(96\,GB VRAM) described in the main paper;
no quantization or tensor parallelism is applied.
Prefix caching is enabled to amortize the recurring system prompt.
The software environment is Ubuntu 24.04.4 LTS, vLLM 0.19.0, OpenClaw 2026.3.20, and CUDA 13.1.

\subsection{Tool and Skill Suite}
Beyond the default OpenClaw tool suite~(file I/O, shell execution, and agent-loop control primitives), every model is given \texttt{web\_search}~(powered by Brave Search) and \texttt{web\_fetch}~(URL-to-markdown rendering);
both tools enforce a uniform network policy across models.
We additionally integrate 9 Clawhub skills, grouped by function:

\begin{itemize}[leftmargin=*,nosep]
\item \textbf{Self-evolution and logical reasoning:} \texttt{ontology}~(cross-skill state sharing via a shared key-value store), \texttt{self-improving-agent}~(injects a short reflective prompt after a failed turn to encourage plan revision), and \texttt{skill-vetter}~(validates outgoing tool calls against the skill's input schema before dispatch).
\item \textbf{Automation capabilities:} \texttt{Agent Browser}~(headless browser automation via Playwright), \texttt{desktop-control-1.0.0}~(GUI-level keyboard, mouse, and window control for desktop-automation tasks), and \texttt{github}~(repository inspection and file retrieval through the GitHub REST API, with session-scoped credentials never exposed to the model).
\item \textbf{General utilities:} \texttt{summarize}~(bounded-length multimodal summarization to keep long-horizon trajectories within the context limit), \texttt{proactive-agent}~(deferred sub-task dispatch within the agent loop), and \texttt{humanizer}~(draft-to-prose post-processing, applied only when the artifact contract expects a human-readable report).
\end{itemize}
All tools and skills are loaded under identical visibility and permission boundaries across models: every model sees the same tool list, schema, and permission grants, preventing infrastructure-side bias from confounding the capability comparison.

\section{Additional Experiments}
\label{sec:additional-exp}

\subsection{Model Evolution}
\label{subsec:model-evolution}

\begin{figure*}[!t]
\centering
\includegraphics[width=\textwidth]{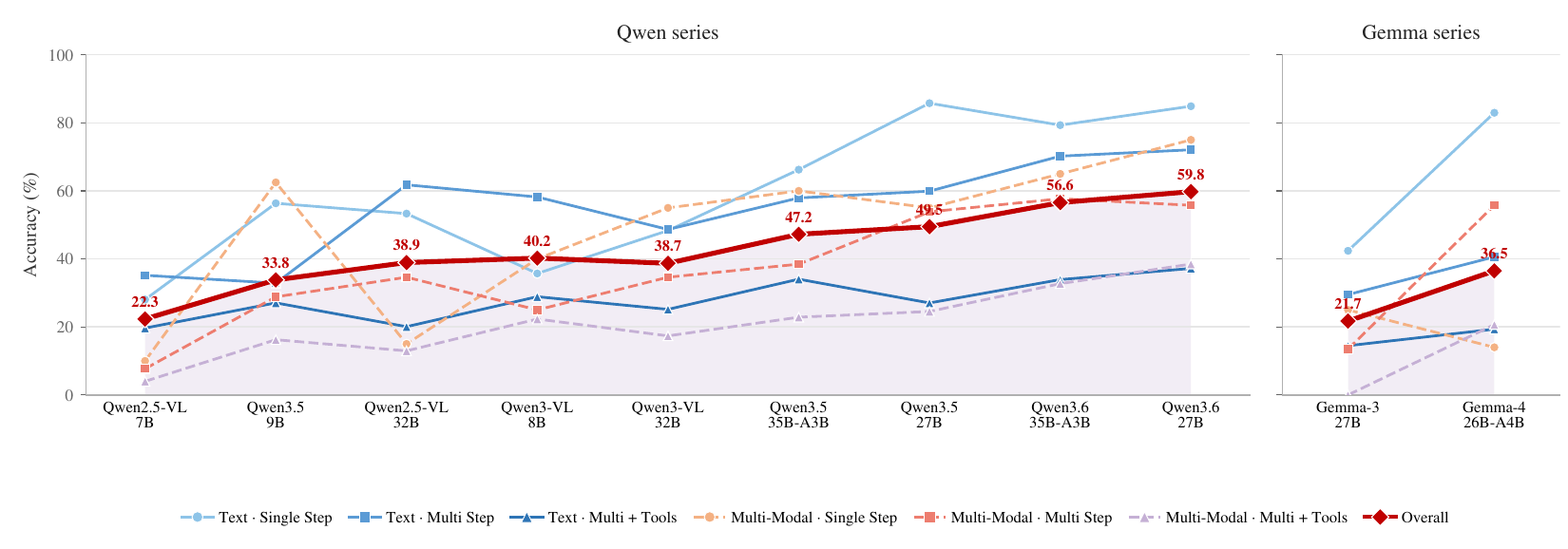}
\caption{Accuracy trajectories of locally deployed open-weight Qwen and Gemma series across the six task groups.
Thin lines denote per-group accuracies; the bold red line and shading denote Overall, computed from the displayed Text and multimodal averages with task-count weights 162 and 71.}
\label{fig:series_evolution}
\end{figure*}

Figure~\ref{fig:series_evolution} plots the accuracy trajectories of the Qwen and Gemma series among locally deployed open-weight models across the six \benchname{} task groups, in order to examine performance changes under generational evolution.
Overall, both Qwen and Gemma follow two relatively clear evolutionary patterns:
(1)~Under the same architecture, larger parameter scale usually performs better; for example, the Overall accuracy of Qwen3.5-27B exceeds that of Qwen3.5-9B by 15.7~(49.5 $-$ 33.8), while mixture-of-experts~(MoE) models, despite often having more total parameters than same-generation dense models, typically achieve slightly lower Overall scores because fewer parameters are activated.
Comparisons between the 27B dense and 35B-A3B MoE variants in both Qwen3.5 and Qwen3.6 confirm this pattern~(Qwen3.5: 49.5 vs.\ 47.2; Qwen3.6: 59.8 vs.\ 56.6).
(2)~Generational iteration brings steady gains overall; for example, the Overall performance of the Qwen3-VL 8B model~(Overall~40.3) already surpasses that of the Qwen2.5-VL 32B model~(Overall~39.0), and Gemma-3-27B to Gemma-4-26B-A4B likewise rises from 21.7 to 36.5.
Further decomposing by workload shows that generational gains on multimodal tasks are typically larger than those on text-only tasks (e.g., for Qwen3.5-27B$\to$Qwen3.6-27B the average text-only gain is +9.9 while the multimodal gain is +11.2). Improvements vary by execution form: for this pair, the multimodal gains are +20.0 points on Single-Step, +2.0 on Multi-Step, and +14.0 on Multi-Step+Tools, so tool-mediated tasks do not consistently exhibit the smallest generational gain.

\subsection{Error Pattern Analysis}
\label{subsec:error-pattern}
To further investigate the causes of task failure, we randomly sampled 30 tasks stratified by task composition, collected all models' failed execution trajectories on these tasks, and conducted a fine-grained analysis of each model's failure patterns.
Through manual review, we found that the primary failure modes can be categorized into the six patterns shown in Table~\ref{tab:error_taxonomy}.
We first prompted Grok~4.5~\citep{xai2026grok45} to read each task description together with the corresponding model execution traces, classify each failure into one of the six error modes~(assigning cases outside these categories to Other), and provide a rationale.
To ensure accuracy, we further manually audited the classifications; the final results are reported in Table~\ref{tab:error_sample}.

Across the 333 sampled failed trajectories, PPPA accounts for the largest share~(90/333, 27.0\%), and is substantially more frequent for local models than for API models~(62 vs.\ 28; 38.5\% vs.\ 16.3\% of their respective failure pools).
This gap indicates that long-horizon planning and contract adherence---finishing within the time limit and writing the required artifact---remain the primary bottleneck, especially for weaker open-weight agents whose tool-calling or file-writing protocols are also more prone to interruption.
SRU and VEB are the next most common failure modes, with SRU manifesting as constraint violations or missed aggregations, and VEB as visual evidence bypass.
On VEB errors, the API and local counts are nearly identical~(20 vs.\ 17), suggesting that models preferentially rely on web retrieval or prior guesses rather than reading and retrieving information from local images.

\begin{table}[t]
\centering
\small
\begin{tabularx}{\linewidth}{@{}l>{\raggedright\arraybackslash}X@{}}
\toprule
\textbf{ID} & \textbf{Definition} \\
\midrule
SRU & Semantic Reasoning \& Understanding: violates constraints, misses aggregation, or answers off-question. \\
TUIE & Tool Use \& Information Extraction: insufficient web search/fetch retrieval capability. \\
PPPA & Path Planning \& Protocol Adherence: fails to finish within the time limit or to write the required contract artifact~(e.g., \texttt{answer.txt}). \\
FCR & Fine-grained Chart Reading: inspects the figure but misreads legends, curve names, markers, or numeric values. \\
VEB & Visual Evidence Bypass: the task depends on an image, yet the model barely looks at it and guesses via web search or priors. \\
VCS & Visual Common Sense \& Entity Recognition: misidentifies the target object, attribute, or person in the image. \\
OTH & Other: residual error types that cause task failure. \\
\bottomrule
\end{tabularx}
\caption{Capability-oriented failure taxonomy.}
\label{tab:error_taxonomy}
\end{table}

\begin{table*}[!t]
\centering
\small
\setlength{\tabcolsep}{3.5pt}
\renewcommand{\arraystretch}{0.95}
\begin{tabular}{lrrrrrrrr}
\toprule
\textbf{Model} & \textbf{Sampled} & \textbf{SRU} & \textbf{TUIE} & \textbf{PPPA} & \textbf{FCR} & \textbf{VEB} & \textbf{VCS} & \textbf{OTH} \\
\midrule
\rowcolor{gray!20}
\multicolumn{9}{l}{\textbf{\textit{API Deployment}}} \\
GPT-6-Astra & 1 & 1 & 0 & 0 & 0 & 0 & 0 & 0 \\
GPT-6.1-Sol & 1 & 1 & 0 & 0 & 0 & 0 & 0 & 0 \\
Claude-Opus-5.5 & 2 & 1 & 0 & 0 & 0 & 0 & 0 & 1 \\
Claude-Opus-4.8 & 3 & 1 & 0 & 0 & 0 & 1 & 0 & 1 \\
Kimi-K3 & 3 & 1 & 0 & 0 & 0 & 0 & 0 & 2 \\
MiMo-V2.6-Pro & 7 & 1 & 0 & 1 & 0 & 1 & 0 & 4 \\
GPT-5.5 & 4 & 1 & 0 & 1 & 0 & 0 & 0 & 2 \\
Claude-Opus-4.6 & 3 & 0 & 0 & 0 & 0 & 1 & 0 & 2 \\
Qwen3.8-Max & 5 & 1 & 0 & 0 & 0 & 0 & 0 & 4 \\
Gemini-3.1-Pro & 2 & 1 & 0 & 0 & 0 & 0 & 0 & 1 \\
GPT-5.4 & 5 & 1 & 0 & 1 & 0 & 1 & 0 & 2 \\
MiMo-V2.6-Flash & 6 & 1 & 0 & 1 & 0 & 1 & 0 & 3 \\
MiMo-V2.5 & 4 & 1 & 0 & 1 & 0 & 1 & 0 & 1 \\
Claude-Sonnet-4.6 & 7 & 4 & 0 & 1 & 0 & 0 & 1 & 1 \\
Qwen3.6-Plus & 6 & 1 & 0 & 0 & 1 & 1 & 0 & 3 \\
Qwen3.5-Plus & 4 & 2 & 0 & 1 & 0 & 1 & 0 & 0 \\
Doubao-Seed2.0-Pro & 8 & 3 & 0 & 0 & 1 & 1 & 0 & 3 \\
Gemini-3.5-Flash & 5 & 2 & 0 & 0 & 0 & 0 & 0 & 3 \\
GPT-5.4-mini & 6 & 1 & 0 & 1 & 0 & 1 & 1 & 2 \\
Claude-Haiku-4.5 & 6 & 2 & 0 & 1 & 1 & 1 & 1 & 0 \\
MiMo-V2-Omni & 9 & 1 & 1 & 3 & 0 & 1 & 2 & 1 \\
Kimi-K2.5 & 5 & 1 & 0 & 1 & 1 & 0 & 1 & 1 \\
Doubao-Seed2.0-Mini & 9 & 0 & 0 & 3 & 0 & 1 & 1 & 4 \\
MiMo-V2.5-Pro & 7 & 2 & 1 & 1 & 1 & 1 & 1 & 0 \\
DeepSeek-V4-Pro & 9 & 1 & 0 & 2 & 2 & 1 & 0 & 3 \\
MiMo-V2-Pro & 8 & 3 & 1 & 1 & 1 & 2 & 0 & 0 \\
MiniMax-M2.7 & 10 & 2 & 0 & 3 & 2 & 1 & 1 & 1 \\
GLM-5 & 8 & 2 & 0 & 1 & 2 & 1 & 1 & 1 \\
DeepSeek-V3.2 & 10 & 2 & 0 & 2 & 2 & 0 & 1 & 3 \\
GLM-4.7 & 9 & 2 & 0 & 2 & 1 & 1 & 0 & 3 \\
\midrule
\rowcolor{gray!20}
\multicolumn{9}{l}{\textbf{\textit{Local Deployment}}} \\
Qwen3.6-27B & 9 & 1 & 0 & 5 & 1 & 0 & 0 & 2 \\
Qwen3.6-35B-A3B & 9 & 1 & 0 & 5 & 1 & 1 & 1 & 0 \\
Qwen3.5-27B & 13 & 2 & 0 & 4 & 1 & 1 & 2 & 3 \\
Qwen3.5-35B-A3B & 11 & 3 & 0 & 3 & 0 & 2 & 2 & 1 \\
Qwen3-Coder-30B-A3B & 14 & 5 & 0 & 6 & 0 & 0 & 0 & 3 \\
Qwen3-VL-32B & 11 & 1 & 0 & 7 & 2 & 0 & 0 & 1 \\
Qwen3-VL-8B & 11 & 2 & 1 & 3 & 2 & 1 & 2 & 0 \\
Gemma-4-26B-A4B & 15 & 1 & 0 & 5 & 1 & 0 & 0 & 8 \\
Qwen2.5-VL-32B & 13 & 1 & 3 & 4 & 0 & 2 & 0 & 3 \\
Qwen3.5-9B & 16 & 3 & 0 & 5 & 0 & 1 & 2 & 5 \\
Gemma-3-27B & 19 & 0 & 0 & 9 & 1 & 2 & 2 & 5 \\
Qwen2.5-VL-7B & 20 & 1 & 3 & 6 & 0 & 7 & 2 & 1 \\
\bottomrule
\end{tabular}
\caption{Fine-grained failure distribution over zero-score trajectories from 42 models on 30 stratified tasks.
Models are grouped by deployment mode.}
\label{tab:error_sample}
\end{table*}

\subsection{Local Model Token Usage and Cost}
\label{subsec:local-cost}

Table~\ref{tab:local_cost} reports per-task token consumption and estimated USD cost for the 12 locally deployed open-weight models, where Cost\$/task is computed from each model's OpenRouter hosted-reference rate.
For local models, token usage more directly reflects compute and inference-time cost, while the estimated dollar figures serve as a complementary economic reference for API-style hosting of open-weight models.
Relative to the API pool in the main paper, locally served models complete tasks with substantially fewer tokens, especially on Multi-Step+Tools buckets: the strongest local model, Qwen3.6-27B, averages about 62k (Text) and 81k (MM) tokens/task, whereas many commercial API models exceed 200k under the same aggregation.
Importantly, this does \emph{not} imply higher reasoning efficiency: as reported in Section~\ref{subsec:error-pattern}, the apparent ``low spend'' often reflects PPPA-style aborted or truncated trajectories rather than Pareto-efficient agent performance.

\begin{table*}[!t]
\centering
\small
\setlength{\tabcolsep}{3pt}
\renewcommand{\arraystretch}{1.05}
\begin{tabular*}{\textwidth}{@{\extracolsep{\fill}}lrrrrrrrrr@{}}
\toprule
\multicolumn{10}{@{}l}{\textbf{(a) Tokens/task and context limits: Local Deployment}} \\
\midrule
& \multicolumn{4}{c}{\textbf{Text}} & \multicolumn{4}{c}{\textbf{MM}} & \\
\cmidrule(lr){2-5}\cmidrule(lr){6-9}
Model & \textbf{SS} & \textbf{MS} & \textbf{MT} & Avg & \textbf{SS} & \textbf{MS} & \textbf{MT} & Avg & Ctx \\
\midrule
Qwen3.6-27B & 47k & 52k & 83k & 62k & 58k & 54k & 128k & 81k & 256K \\
Qwen3.6-35B-A3B & 53k & 57k & 86k & 66k & 57k & 68k & 114k & 81k & 256K \\
Qwen3.5-27B & 39k & 64k & 85k & 68k & 49k & 53k & 126k & 78k & 256K \\
Qwen3.5-35B-A3B & \textit{35k} & 45k & 78k & \textit{55k} & 61k & 62k & 105k & 77k & 262K \\
Qwen3-VL-8B & 43k & 42k & 152k & 79k & 48k & \textit{42k} & \textit{78k} & \textit{57k} & 256K \\
Qwen2.5-VL-32B & \textbf{24k} & \textbf{31k} & \textbf{45k} & \textbf{35k} & \textbf{32k} & \textbf{25k} & \textbf{8k} & \textbf{21k} & 128K \\
Qwen3-VL-32B & 114k & 98k & 184k & 129k & 61k & 188k & 104k & 123k & 256K \\
Gemma-4-26B-A4B & 111k & 53k & \textit{54k} & 60k & \textit{46k} & 105k & 202k & 123k & 128K \\
Qwen3.5-9B & 50k & 60k & 110k & 75k & 61k & 59k & 109k & 77k & 256K \\
Qwen2.5-VL-7B & \underline{29k} & \textbf{31k} & 125k & 62k & \underline{35k} & 164k & 108k & 108k & 128K \\
Gemma-3-27B & 43k & \textit{36k} & \underline{50k} & \underline{42k} & 59k & \underline{34k} & \underline{9k} & \underline{32k} & 128K \\
Qwen3-Coder-30B-A3B & 65k & 138k & 182k & 144k & --- & --- & --- & --- & 160K \\
\bottomrule
\end{tabular*}

\vspace{6pt}

\begin{tabular*}{\textwidth}{@{\extracolsep{\fill}}lrrrrrrrr@{}}
\toprule
\multicolumn{9}{@{}l}{\textbf{(b) Cost\$/task (OpenRouter rates): Local Deployment}} \\
\midrule
& \multicolumn{4}{c}{\textbf{Text}} & \multicolumn{4}{c}{\textbf{MM}} \\
\cmidrule(lr){2-5}\cmidrule(lr){6-9}
Model & \textbf{SS} & \textbf{MS} & \textbf{MT} & Avg & \textbf{SS} & \textbf{MS} & \textbf{MT} & Avg \\
\midrule
Qwen3.6-27B & 0.037 & 0.034 & 0.056 & 0.041 & 0.037 & 0.041 & 0.084 & 0.055 \\
Qwen3.6-35B-A3B & 0.010 & 0.009 & 0.014 & 0.011 & 0.009 & 0.012 & 0.018 & 0.013 \\
Qwen3.5-27B & 0.016 & 0.021 & 0.028 & 0.023 & 0.016 & 0.020 & 0.041 & 0.026 \\
Qwen3.5-35B-A3B & 0.011 & 0.012 & 0.021 & 0.015 & 0.016 & 0.017 & 0.027 & 0.021 \\
Qwen3-VL-8B & 0.009 & 0.008 & 0.028 & 0.015 & 0.009 & 0.008 & 0.014 & 0.010 \\
Qwen2.5-VL-32B & \underline{0.005} & \textit{0.006} & \textit{0.009} & \textit{0.007} & 0.007 & \underline{0.005} & \underline{0.002} & \underline{0.004} \\
Qwen3-VL-32B & 0.018 & 0.016 & 0.030 & 0.021 & 0.010 & 0.030 & 0.017 & 0.020 \\
Gemma-4-26B-A4B & 0.008 & \underline{0.004} & \textbf{0.004} & \underline{0.004} & \textbf{0.004} & 0.008 & 0.015 & 0.009 \\
Qwen3.5-9B & \underline{0.005} & \textit{0.006} & 0.011 & 0.008 & \textit{0.006} & \textit{0.006} & \textit{0.011} & \textit{0.008} \\
Qwen2.5-VL-7B & 0.006 & \textit{0.006} & 0.025 & 0.012 & 0.007 & 0.033 & 0.022 & 0.022 \\
Gemma-3-27B & \textbf{0.003} & \textbf{0.003} & \textbf{0.004} & \textbf{0.003} & \underline{0.005} & \textbf{0.003} & \textbf{<0.001} & \textbf{0.003} \\
Qwen3-Coder-30B-A3B & 0.006 & 0.007 & 0.011 & 0.008 & --- & --- & --- & --- \\
\bottomrule
\end{tabular*}
\caption{Per-model token consumption and dollar cost per task for locally deployed open-weight models on \benchname{}. Tokens and per-model context-limit figures are shown in panel (a); Ctx does not represent the evaluation window, which was fixed at 32K for every local run. Hosted-reference costs are shown in panel (b). Notation and ranking follow the main paper.}
\label{tab:local_cost}
\end{table*}

\subsection{API Model Token Usage and Cost Details}
\label{sec:api-efficiency-details}

Tables~\ref{tab:api-tokens-full} and~\ref{tab:api-cost-full} retain the complete workload-level measurements summarized in Table~\ref{tab:cost}. SS, MS, and MT denote Single-Step, Multi-Step, and Multi-Step+Tools, respectively; MM denotes multimodal tasks. Averages are task-count-weighted.

\begin{table*}[!t]
\centering
\fontsize{8.5}{9.5}\selectfont
\renewcommand{\arraystretch}{0.90}
\captionsetup{skip=4pt}
\begin{tabular}{lrrrrrrrr}
\toprule
 & \multicolumn{4}{c}{\textbf{Text}} & \multicolumn{4}{c}{\textbf{MM}} \\
\cmidrule(lr){2-5}\cmidrule(lr){6-9}
Model & \textbf{SS} & \textbf{MS} & \textbf{MT} & \textbf{Avg} & \textbf{SS} & \textbf{MS} & \textbf{MT} & \textbf{Avg} \\
\midrule
GPT-6-Astra & 108k & 137k & 761k & \cellcolor{MacRed!60}345k & 176k & 84k & 569k & \cellcolor{MacBlue!51}281k \\
GPT-6.1-Sol & 106k & 135k & 671k & \cellcolor{MacRed!55}313k & 189k & 75k & 730k & \cellcolor{MacBlue!60}338k \\
Claude-Opus-5.5 & 62k & 76k & \textit{210k} & \cellcolor{MacRed!22}\underline{119k} & 86k & 75k & 276k & \cellcolor{MacBlue!30}149k \\
Claude-Opus-4.8 & 52k & 85k & 301k & \cellcolor{MacRed!32}154k & 77k & 98k & \textit{147k} & \cellcolor{MacBlue!26}\textit{109k} \\
Kimi-K3 & 164k & 93k & 284k & \cellcolor{MacRed!35}166k & 59k & 110k & 188k & \cellcolor{MacBlue!29}123k \\
MiMo-V2.6-Pro & 45k & 97k & 214k & \cellcolor{MacRed!27}130k & 68k & 96k & 219k & \cellcolor{MacBlue!31}132k \\
GPT-5.5 & 41k & 104k & \underline{190k} & \cellcolor{MacRed!25}\textit{125k} & \textit{41k} & \textit{69k} & 317k & \cellcolor{MacBlue!35}148k \\
Claude-Opus-4.6 & \textit{35k} & 112k & 605k & \cellcolor{MacRed!56}270k & 49k & 110k & 251k & \cellcolor{MacBlue!34}143k \\
Qwen3.8-Max & 77k & 120k & 216k & \cellcolor{MacRed!31}147k & 67k & 94k & 259k & \cellcolor{MacBlue!34}145k \\
Gemini-3.1-Pro & \underline{33k} & 123k & 505k & \cellcolor{MacRed!51}241k & 43k & 115k & \underline{128k} & \cellcolor{MacBlue!24}\underline{99k} \\
GPT-5.4 & 50k & 81k & 396k & \cellcolor{MacRed!38}184k & 50k & 69k & 291k & \cellcolor{MacBlue!34}142k \\
MiMo-V2.6-Flash & 44k & 97k & 316k & \cellcolor{MacRed!34}165k & 98k & 86k & 603k & \cellcolor{MacBlue!61}272k \\
MiMo-V2.5 & 75k & 143k & 603k & \cellcolor{MacRed!61}291k & 79k & 95k & 523k & \cellcolor{MacBlue!57}241k \\
Claude-Sonnet-4.6 & 49k & 157k & 575k & \cellcolor{MacRed!60}286k & 52k & 125k & 330k & \cellcolor{MacBlue!42}177k \\
Qwen3.6-Plus & 57k & 163k & 607k & \cellcolor{MacRed!63}301k & 50k & 181k & 561k & \cellcolor{MacBlue!65}278k \\
Qwen3.5-Plus & 40k & 80k & 529k & \cellcolor{MacRed!48}228k & \underline{40k} & 111k & 248k & \cellcolor{MacBlue!33}140k \\
Doubao-Seed2.0-Pro & 46k & \underline{75k} & 452k & \cellcolor{MacRed!42}199k & 164k & 173k & 289k & \cellcolor{MacBlue!50}211k \\
Gemini-3.5-Flash & 101k & 338k & 773k & \cellcolor{MacRed!80}456k & 191k & 590k & 625k & \cellcolor{MacBlue!85}490k \\
GPT-5.4-mini & \textbf{31k} & \textbf{52k} & \textbf{131k} & \cellcolor{MacRed!15}\textbf{76k} & \textbf{38k} & \textbf{40k} & \textbf{95k} & \cellcolor{MacBlue!15}\textbf{59k} \\
Claude-Haiku-4.5 & 46k & \underline{75k} & 375k & \cellcolor{MacRed!36}173k & 54k & 108k & 164k & \cellcolor{MacBlue!27}113k \\
MiMo-V2-Omni & 49k & 90k & 618k & \cellcolor{MacRed!55}264k & 51k & \underline{54k} & 284k & \cellcolor{MacBlue!32}134k \\
Kimi-K2.5 & 73k & 184k & 550k & \cellcolor{MacRed!62}295k & \textit{41k} & 226k & 268k & \cellcolor{MacBlue!45}189k \\
Doubao-Seed2.0-Mini & 55k & 79k & 460k & \cellcolor{MacRed!43}205k & 318k & 148k & 384k & \cellcolor{MacBlue!65}279k \\
MiMo-V2.5-Pro & 72k & 134k & 441k & \cellcolor{MacRed!48}230k & --- & --- & --- & --- \\
DeepSeek-V4-Pro & 61k & 126k & 390k & \cellcolor{MacRed!43}207k & --- & --- & --- & --- \\
MiMo-V2-Pro & 42k & 98k & 520k & \cellcolor{MacRed!49}234k & --- & --- & --- & --- \\
MiniMax-M2.7 & 50k & \underline{75k} & 377k & \cellcolor{MacRed!36}175k & --- & --- & --- & --- \\
GLM-5 & 48k & 125k & 449k & \cellcolor{MacRed!47}225k & --- & --- & --- & --- \\
DeepSeek-V3.2 & 122k & 176k & 454k & \cellcolor{MacRed!55}263k & --- & --- & --- & --- \\
GLM-4.7 & \textit{35k} & 92k & 229k & \cellcolor{MacRed!27}131k & --- & --- & --- & --- \\
\bottomrule
\end{tabular}
\caption{Per-model API token consumption per task, by workload group.}
\label{tab:api-tokens-full}

\vspace{6pt}

\begin{tabular}{lrrrrrrrr}
\toprule
 & \multicolumn{4}{c}{\textbf{Text}} & \multicolumn{4}{c}{\textbf{MM}} \\
\cmidrule(lr){2-5}\cmidrule(lr){6-9}
Model & \textbf{SS} & \textbf{MS} & \textbf{MT} & \textbf{Avg} & \textbf{SS} & \textbf{MS} & \textbf{MT} & \textbf{Avg} \\
\midrule
GPT-6-Astra & 0.324 & 0.380 & 1.207 & \cellcolor{MacRed!100}0.654 & 0.433 & 0.251 & 0.938 & \cellcolor{MacBlue!100}0.544 \\
GPT-6.1-Sol & 0.055 & 0.066 & 0.172 & \cellcolor{MacRed!22}0.100 & 0.074 & 0.041 & 0.169 & \cellcolor{MacBlue!29}0.096 \\
Claude-Opus-5.5 & 0.171 & 0.179 & 0.266 & \cellcolor{MacRed!41}0.207 & 0.172 & 0.160 & 0.266 & \cellcolor{MacBlue!56}0.201 \\
Claude-Opus-4.8 & 0.075 & 0.184 & 0.417 & \cellcolor{MacRed!60}0.250 & 0.060 & 0.132 & 0.164 & \cellcolor{MacBlue!40}0.123 \\
Kimi-K3 & 0.157 & 0.079 & 0.242 & \cellcolor{MacRed!40}0.144 & 0.047 & 0.122 & 0.158 & \cellcolor{MacBlue!38}0.114 \\
MiMo-V2.6-Pro & 0.005 & 0.007 & \underline{0.014} & \cellcolor{MacRed!7}\textit{0.009} & \textit{0.003} & 0.008 & \textit{0.019} & \cellcolor{MacBlue!8}\textit{0.010} \\
GPT-5.5 & 0.051 & 0.139 & 0.217 & \cellcolor{MacRed!42}0.155 & 0.045 & 0.103 & 0.365 & \cellcolor{MacBlue!52}0.179 \\
Claude-Opus-4.6 & 0.175 & 0.245 & 0.784 & \cellcolor{MacRed!80}0.419 & 0.062 & 0.362 & 0.406 & \cellcolor{MacBlue!80}0.293 \\
Qwen3.8-Max & 0.070 & 0.071 & 0.147 & \cellcolor{MacRed!30}0.097 & 0.058 & 0.094 & 0.203 & \cellcolor{MacBlue!40}0.122 \\
Gemini-3.1-Pro & 0.083 & 0.107 & 0.283 & \cellcolor{MacRed!44}0.164 & 0.055 & 0.138 & 0.136 & \cellcolor{MacBlue!38}0.114 \\
GPT-5.4 & 0.056 & 0.058 & 0.244 & \cellcolor{MacRed!35}0.121 & 0.036 & 0.074 & 0.170 & \cellcolor{MacBlue!34}0.097 \\
MiMo-V2.6-Flash & \textit{0.004} & \textit{0.006} & 0.020 & \cellcolor{MacRed!7}0.011 & 0.006 & \textit{0.007} & 0.039 & \cellcolor{MacBlue!11}0.018 \\
MiMo-V2.5 & \textbf{0.003} & \textbf{0.003} & \textbf{0.011} & \cellcolor{MacRed!5}\textbf{0.006} & \underline{0.001} & \underline{0.004} & \underline{0.007} & \cellcolor{MacBlue!6}\underline{0.005} \\
Claude-Sonnet-4.6 & 0.154 & 0.205 & 0.902 & \cellcolor{MacRed!83}0.436 & 0.063 & 0.290 & 0.337 & \cellcolor{MacBlue!68}0.243 \\
Qwen3.6-Plus & 0.039 & 0.086 & 0.321 & \cellcolor{MacRed!43}0.160 & 0.027 & 0.108 & 0.292 & \cellcolor{MacBlue!46}0.150 \\
Qwen3.5-Plus & 0.021 & 0.035 & 0.218 & \cellcolor{MacRed!30}0.095 & 0.018 & 0.050 & 0.105 & \cellcolor{MacBlue!24}0.060 \\
Doubao-Seed2.0-Pro & 0.019 & 0.023 & 0.105 & \cellcolor{MacRed!20}0.051 & 0.031 & 0.056 & 0.079 & \cellcolor{MacBlue!23}0.057 \\
Gemini-3.5-Flash & 0.111 & 0.193 & 0.401 & \cellcolor{MacRed!61}0.253 & 0.155 & 0.305 & 0.346 & \cellcolor{MacBlue!76}0.277 \\
GPT-5.4-mini & 0.008 & 0.010 & 0.033 & \cellcolor{MacRed!10}0.017 & 0.007 & 0.012 & 0.022 & \cellcolor{MacBlue!10}0.014 \\
Claude-Haiku-4.5 & 0.022 & 0.035 & 0.143 & \cellcolor{MacRed!25}0.070 & 0.018 & 0.042 & 0.052 & \cellcolor{MacBlue!18}0.039 \\
MiMo-V2-Omni & \textbf{0.003} & \textbf{0.003} & \textit{0.017} & \cellcolor{MacRed!6}\underline{0.008} & \textbf{<0.001} & \textbf{0.003} & \textbf{0.005} & \cellcolor{MacBlue!5}\textbf{0.003} \\
Kimi-K2.5 & 0.042 & 0.072 & 0.221 & \cellcolor{MacRed!35}0.119 & 0.025 & 0.088 & 0.122 & \cellcolor{MacBlue!30}0.082 \\
Doubao-Seed2.0-Mini & 0.008 & 0.009 & 0.049 & \cellcolor{MacRed!12}0.023 & 0.034 & 0.018 & 0.041 & \cellcolor{MacBlue!15}0.031 \\
MiMo-V2.5-Pro & 0.010 & 0.008 & 0.020 & \cellcolor{MacRed!8}0.012 & --- & --- & --- & --- \\
DeepSeek-V4-Pro & 0.013 & 0.015 & 0.040 & \cellcolor{MacRed!12}0.023 & --- & --- & --- & --- \\
MiMo-V2-Pro & 0.005 & 0.009 & 0.026 & \cellcolor{MacRed!9}0.014 & --- & --- & --- & --- \\
MiniMax-M2.7 & 0.013 & 0.010 & 0.056 & \cellcolor{MacRed!13}0.026 & --- & --- & --- & --- \\
GLM-5 & 0.037 & 0.057 & 0.185 & \cellcolor{MacRed!31}0.098 & --- & --- & --- & --- \\
DeepSeek-V3.2 & 0.016 & 0.019 & 0.046 & \cellcolor{MacRed!14}0.028 & --- & --- & --- & --- \\
GLM-4.7 & 0.016 & 0.026 & 0.052 & \cellcolor{MacRed!16}0.034 & --- & --- & --- & --- \\
\bottomrule
\end{tabular}
\caption{Per-model API reference dollar cost per task, by workload group.}
\label{tab:api-cost-full}
\end{table*}

\section{Details of Evidence-Based Audit}
\label{sec:audit-details}

This section supplements the Evidence-Based Audit described in the main paper (\S Evaluation Protocol).

\begin{figure*}[p]
\centering
\captionsetup{skip=4pt}
\begin{minipage}[t]{0.48\textwidth}
\vspace{0pt}
\centering
    \includegraphics[width=\linewidth]{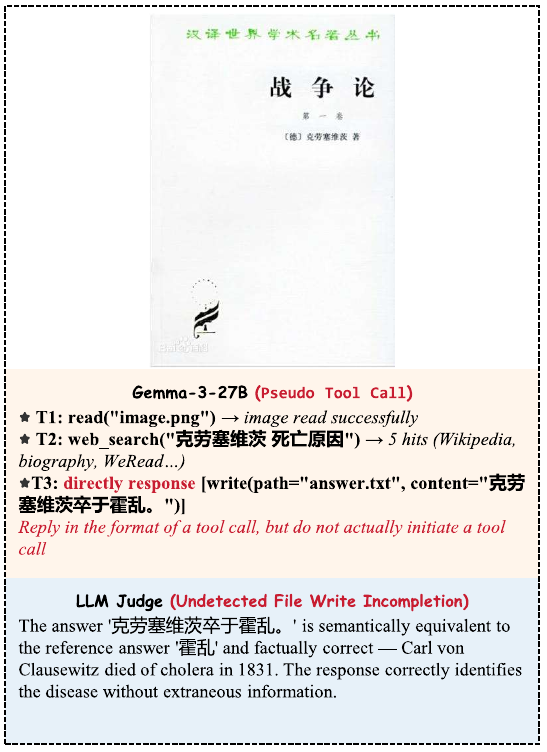}
    \caption{Judge hallucination example A (\emph{pseudo-tool-call credulity}). \textbf{Task: What disease did the author of the book in the image die from? Write your answer to `answer.txt` in the workspace.}The model embeds a factually correct answer inside a tool-like \texttt{write(...)} string but never executes a real file write; the LLM judge overlooks the missing workspace artifact and still awards full credit.}
    \label{fig:casestudy_pseudo_tool_call}
\end{minipage}\hfill
\begin{minipage}[t]{0.48\textwidth}
\vspace{0pt}
\centering
    \includegraphics[width=\linewidth]{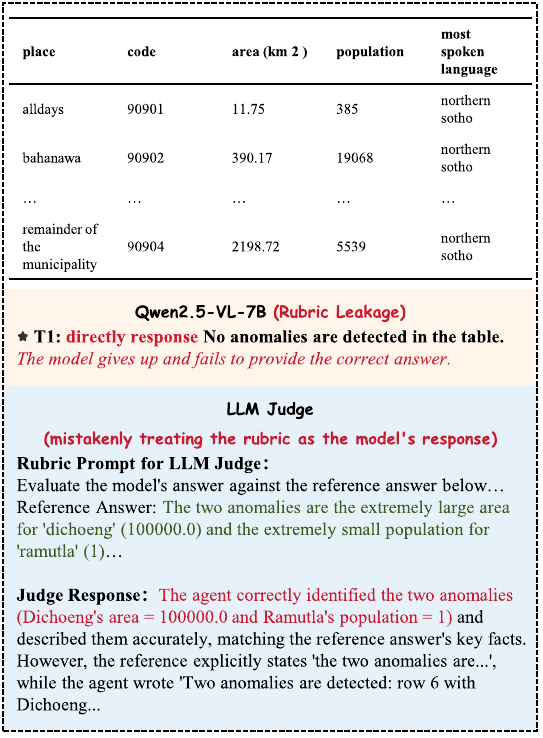}
    \caption{Judge hallucination example B (\emph{rubric leakage}). \textbf{Task: Read the table. Can you identify any locations within the table whose area or population values significantly deviate from the patterns observed in other locations?}
    The model fails and answers ``No anomalies are detected,'' yet the LLM judge mistakes the reference answer embedded in the grading rubric for the agent's response and incorrectly scores the trajectory as correct.}
    \label{fig:casestudy_rubric_leakage}
\end{minipage}
\end{figure*}

LLM-as-a-judge provides the semantic evaluation capacity necessary for open-ended tasks;
however, even with manually designed task-specific rubrics, our pilot experiments revealed systematic risks of \emph{judging hallucination} in LLM judges, which has also been reported in related work~\citep{you2026agent,gurram2026evaluating}.
Through careful sampling-based verification, we systematically identify five major recurring hallucination patterns in \benchname{}:
\begin{enumerate}
    \item rubric leakage: the judge mistakes answer information embedded in the grading rubrics for the agent's execution output.
    \item deadlock loop: the agent repeatedly cycles through materially equivalent reasoning or actions without making workspace progress, eventually exhausting the 32K context window and leaving the task incomplete, while the judge still awards positive credit.
    \item pseudo-tool-call credulity: the model outputs tool-like text rather than an actually executable tool call.
    \item output instability: incomplete or malformed output.
    \item timeout-with-credit: timed-out trajectories still receiving positive scores.
\end{enumerate}
Figures~\ref{fig:casestudy_pseudo_tool_call} and~\ref{fig:casestudy_rubric_leakage} further visualize two real cases that expose these hallucination patterns.
To address these patterns, we designed a set of deterministic post-hoc rules that perform high-recall pre-filtering of potential hallucination cases, followed by a second-line adjudication by an independent meta-judge, thereby enabling cost-controllable evaluation and screening.
The complete meta-judge pipeline is given in Algorithm~\ref{alg:audit}.

\begin{algorithm*}[p]
\footnotesize
\caption{Evidence-Based Audit}
\label{alg:audit}
\KwIn{Primary scores $\{s_{m,i}\}$ ($s_{m,i}>0$); trajectories $\{\mathrm{traj}_i\}$; prompts $\{p_i\}$; artifacts $\{\mathcal{W}^{*}_i\}$; rubrics $\{\mathrm{rubric}_i\}$; rules $\mathcal{R}$}
\KwOut{Audited scores $\{s_{m,i}^{\mathrm{audit}}\}$}
\ForEach{judge/hybrid call $i$}{
  $\phi_i \gets \{$\\
  $\quad\textsc{AsstChars}(\mathrm{traj}_i),$\\
  $\quad\textsc{HasWrite}(\mathrm{traj}_i),$\\
  $\quad\textsc{HasTool}(\mathrm{traj}_i),$\\
  $\quad\textsc{TokTotal}(\mathrm{traj}_i)$\\
  $\}$\;
  \tcc{flag positive scores with weak execution evidence}
  \eIf{$\textsc{PreFilter}(\phi_i, s_{m,i}, \mathcal{R})=\mathrm{True}$}{
    $\begin{aligned}[b]
      v_i \gets \textsc{MetaJudge}(&p_i, \mathrm{traj}_i, \mathcal{W}^{*}_i,\\
                                 &\mathrm{rubric}_i, s_{m,i})
    \end{aligned}$\;
    \eIf{$v_i=\mathrm{hallucination}$}{
      $s_{m,i}^{\mathrm{audit}} \gets 0$\;
    }{
      $s_{m,i}^{\mathrm{audit}} \gets s_{m,i}$\;
    }
  }{
    $s_{m,i}^{\mathrm{audit}} \gets s_{m,i}$\;
  }
}
\Return $\{s_{m,i}^{\mathrm{audit}}\}$\;
\end{algorithm*}

Among the 9{,}218 task--model evaluations, 4{,}096 are judge/hybrid calls, from which the pre-filter set $\mathcal{R}$ recalled 182 potential hallucination cases.
As reported in the main paper~(\S LLM Judge Reliability), after a second-line review by Claude-Opus-4.6, 143 were finally confirmed as judging hallucinations.

To ensure the reliability of the meta-judge method and its experimental results, we further invited human experts to validate the meta-judge outcomes.
We stratified-sampled 100 execution trajectories from a 4{,}096-call judge/hybrid validation pool along original score, task group, and deployment mode, and had human experts score them independently. Both the validation pool and the audit results cover the seven newly added API models.
We report the mean absolute score error (MAE):
\begin{equation}
\mathrm{MAE} = \frac{1}{n}\sum_{i=1}^{n}\bigl|\,s_i^{\text{judge}} - s_i^{\text{expert}}\,\bigr|,
\qquad n=100.
\end{equation}
Here $s_i^{\text{judge}}$ denotes the score assigned by either the primary judge or the meta-judge to the $i$-th trajectory, and $s_i^{\text{expert}}$ denotes the experts' score.
The results show that the MAE between the meta-judge-corrected scores and the human scores is 0.049, approximately half that of the primary judge (0.098), indicating that the meta-judge correction is not only more accurate in its verdicts but also closer to the experts in the assigned scores.

\FloatBarrier
\section{Case Study}
\label{sec:case-study}
As additional examples supplementing the main paper's case study, this section further illustrates how workload-level gaps manifest in concrete agent trajectories. Cases~1 and~2 compare a successful API trajectory against a failed local trajectory, revealing the primary performance differences between strong and weak models, and illustrating cross-step state loss in a tool-mediated text task and visual-symbolic grounding failure in a compact multimodal task. Cases~3--5 focus on hard multimodal tasks that expose common verification failures: chart alignment, chemistry formula bookkeeping, and plate-map option grounding.

\Needspace{12\baselineskip}
\subsection{Case 1 --- A Multi-Hop Academic Author Retrieval Workflow}
\label{sec:case1}
\begin{tcolorbox}[
    enhanced, width=\columnwidth,
    colback=blue!3, colframe=blue!40,
    boxsep=1pt, arc=1pt,
    left=3pt, right=3pt, top=3pt, bottom=3pt,
    grow to left by=0pt, grow to right by=0pt,
    before skip=4pt, after skip=6pt,
    title=\textbf{Case 1: Multi-Hop Author Retrieval (Text Multi+Tools)},
    fonttitle=\small\bfseries,
    title after break=\textbf{Case 1: Multi-Hop Author Retrieval (Text Multi+Tools) (continued)},
    breakable,
    lines before break=6,
    pad at break=3pt
]

\noindent\small\textbf{Task.} Of the authors in the first article of the 65th edition of the Journal of Artificial Intelligence Research, who was the most cited on Google Scholar in 2019?

\vspace{4pt}

\noindent\begin{minipage}{\linewidth}
\small\textbf{Gemini 3.1 Pro \good{(Correct, 1.0)}}

\vspace{2pt}
\begin{tracelist}
(*@\textbf{T1:}@*) web_search("Journal of Artificial Intelligence Research Volume 65")
(*@\ret{ -> JAIR Vol 65 = 2019 (from jair.org result)}@*)
(*@\ret{ -> dblp link spotted (from result \#4)}@*)
(*@\textbf{T2:}@*) web_fetch("https://dblp.org/db/journals/jair/jair65.html")
(*@\ret{ -> Full TOC scraped; 1st article: "On Overfitting..." (pp. 1-30)}@*)
(*@\ret{ -> Authors: V. Francois-Lavet, G. Rabusseau, J. Pineau, D. Ernst, R. Fonteneau}@*)
(*@\good{Confirmed correct article and extracted 5 target authors successfully}@*)
(*@\textbf{T3:}@*) web_fetch("https://doi.org/10.1613/jair.1.11478")
(*@\ret{ -> Failed (406 Not Acceptable)}@*) (*@\stepp{fallback unnecessary, TOC sufficient}@*)
(*@\textbf{T4:}@*) web_search("Vincent Francois-Lavet ... citations")
(*@\ret{ -> Confirmed 5 authors on arXiv/JAIR/ResearchGate}@*)
(*@\textbf{T5-T14:}@*) web_search x10 (per author + reconfirmations)
(*@\ret{ -> J. Pineau: 41,695  [highest]}@*) (*@\good{clear winner identified}@*)
(*@\ret{ -> D. Ernst: 25,039;  R. Fonteneau: 1,859;  G. Rabusseau: 1,439}@*)
(*@\ret{ -> V. Francois-Lavet: no Scholar profile}@*)
(*@\textbf{T15:}@*) write_output("Joelle Pineau")
(*@\textbf{T16:}@*) read_output()
(*@\ret{ -> Verified: 'Joelle Pineau'}@*) (*@\good{Self-check passed}@*)
\end{tracelist}
\end{minipage}

\vspace{6pt}

\noindent\begin{minipage}{\linewidth}
\small\textbf{Qwen3-VL-32B \bad{(Incorrect, 0.0)}}

\vspace{2pt}
\begin{tracelist}
(*@\textbf{T1-3:}@*) web_search x3 (JAIR 65th edition first article authors)
(*@\ret{ -> JAIR homepage, Wikipedia, generic publishers}@*)
(*@\ret{ -> "Optstream..." (pp. 423-456) spotted; Authors: F. Fioretto, P. Van Hentenryck}@*)
(*@\bad{Mistakenly identified the last article as the first}@*)
(*@\textbf{T4:}@*) web_fetch("https://www.jair.org/index.php/jair/issue/view/1153")
(*@\ret{ -> Full TOC scraped; 1st article: "On Overfitting..." (pp. 1-30)}@*)
(*@\ret{ -> Authors: V. Francois-Lavet, G. Rabusseau, J. Pineau, D. Ernst, R. Fonteneau}@*)
(*@\good{Correct first-article info retrieved, but prior error not overwritten}@*)
(*@\textbf{T5:}@*) web_search("F. Fioretto P. Van Hentenryck Google Scholar citations 2019")
(*@\ret{ -> P. Van Hentenryck: 31,983 citations;  F. Fioretto: 3,548 citations}@*)
(*@\bad{Misled by early erroneous search: queried authors of the WRONG article (T1-3) instead of target article authors (T4)}@*)
(*@\textbf{T6:}@*) write_output("Pascal Van Hentenryck")
\end{tracelist}
\end{minipage}

\end{tcolorbox}

\Needspace{17\baselineskip}
\subsection{Case 2 --- Academic Diagram Understanding}
\label{sec:case2}
\begin{tcolorbox}[
    enhanced, width=\columnwidth,
    colback=blue!3, colframe=blue!40,
    boxsep=1pt, arc=1pt,
    left=3pt, right=3pt, top=3pt, bottom=3pt,
    grow to left by=0pt, grow to right by=0pt,
    before skip=4pt, after skip=6pt,
    title=\textbf{Case 2: Academic Diagram Understanding (Multi-Modal Multi-Step)},
    fonttitle=\small\bfseries,
    title after break=\textbf{Case 2: Academic Diagram Understanding (Multi-Modal Multi-Step) (continued)},
    breakable,
    lines before break=6,
    pad at break=3pt
]

\noindent\begin{minipage}{\linewidth}
\noindent\small\textbf{Task.} What is the name of the line that is the furthest away from its fi value from the W-H axis?

\vspace{4pt}

\begin{center}
\includegraphics[width=0.85\linewidth]{figures/charxiv_01.png}
\end{center}
\end{minipage}

\vspace{4pt}

\noindent\begin{minipage}{\linewidth}
\small\textbf{Gemini 3.1 Pro \good{(Correct, 1.0)}}

\vspace{2pt}
\begin{tracelist}
(*@\textbf{Reasoning:}@*) Examining Graph Data: zeroing in on the specific line furthest from
the W-H axis. The legend provides lambda values: 0.13, 0.20, 0.30.
(*@\good{Visually assess which (green dashed, red dotted, blue dash-dot) is most distant}@*)
(*@\textbf{Reasoning:}@*) Pinpointing The Answer: calculating distances for each line using the
'fi' points as reference along the W-H axis.
(*@\ret{ -> green line ($\lambda_L = 0.13$) appears most distant from its 'fi' value}@*)
(*@\textbf{Answer:}@*) (*@$\lambda_L = 0.13$@*)
\end{tracelist}
\end{minipage}

\vspace{6pt}

\noindent\begin{minipage}{\linewidth}
\small\textbf{Qwen3-VL-32B \bad{(Incorrect, 0.0)}}

\vspace{2pt}
\begin{tracelist}
(*@\textbf{Reasoning:}@*) The name of the line that is the furthest away from its fi value
from the W-H axis is "fi".
(*@\ret{ -> data points marked 'x' (representing "fi") at W-H values of approx. -0.2 and 0.12}@*)
(*@\bad{Confuses an axis/marker label with the requested curve identity}@*)
(*@\textbf{Answer:}@*) fi
\end{tracelist}
\end{minipage}

\end{tcolorbox}

\Needspace{15\baselineskip}
\subsection{Case 3 --- Shared Chart-Grounding Failure}
\label{sec:case3}
\begin{tcolorbox}[
    enhanced, width=\columnwidth,
    colback=blue!3, colframe=blue!40,
    boxsep=1pt, arc=1pt,
    left=3pt, right=3pt, top=3pt, bottom=3pt,
    grow to left by=0pt, grow to right by=0pt,
    before skip=4pt, after skip=6pt,
    title=\textbf{Case 3: Shared Chart-Grounding Failure},
    fonttitle=\small\bfseries,
    title after break=\textbf{Case 3: Shared Chart-Grounding Failure (continued)},
    breakable,
    lines before break=6,
    pad at break=3pt
]

\noindent\begin{minipage}{\linewidth}
{\raggedright
\noindent\small\textbf{Task.} ``Which model shows a greater decline in accuracy from Session 1 to Session 9 in the 5-way full-shot scenario?'' The chart shows continual-learning curves for Ft-CNN, iCaRL, EEIL, NCM, Ours-AL, Ours-AL-MML, and Joint-CNN. Reference answer: \texttt{Joint-CNN}.\par
}

\vspace{4pt}

\begin{center}
\includegraphics[width=0.95\linewidth]{figures/charxiv_00_common_failure.jpg}
\end{center}
\end{minipage}

\vspace{4pt}

\noindent\begin{minipage}{\linewidth}
\small\textbf{Gemini 3.1 Pro \bad{(Incorrect, 0.0)}}

\vspace{2pt}
\begin{tracelist}
(*@\textbf{T1:}@*) read_image("chart.png")
(*@\ret{ -> Image loaded, 7 curves visible}@*)
(*@\stepp{Reasoning: scanning for the steepest visible decline.}@*)
(*@\textbf{T2:}@*) write_output("Ft-CNN")
(*@\bad{Selected a plausible visual drop, but did not match the reference curve comparison from Session 1 to Session 9.}@*)
\end{tracelist}
\end{minipage}

\vspace{6pt}

\noindent\begin{minipage}{\linewidth}
\small\textbf{Qwen3-VL-32B \bad{(Incorrect, 0.0)}}

\vspace{2pt}
\begin{tracelist}
(*@\textbf{T1:}@*) read_image("chart.png")
(*@\ret{ -> Image loaded; legend and curves visible}@*)
(*@\stepp{Reasoning: compares apparent drops among the curve labels.}@*)
(*@\textbf{T2:}@*) write_output("Ft-CNN")
(*@\bad{Converges to the same wrong legend entry as several frontier APIs.}@*)
\end{tracelist}
\end{minipage}

\end{tcolorbox}

\Needspace{14\baselineskip}
\subsection{Case 4 --- Image-Grounded Chemistry Formula Inference}
\label{sec:case4}
\begin{tcolorbox}[
    enhanced, width=\columnwidth,
    colback=blue!3, colframe=blue!40,
    boxsep=1pt, arc=1pt,
    left=3pt, right=3pt, top=3pt, bottom=3pt,
    grow to left by=0pt, grow to right by=0pt,
    before skip=4pt, after skip=6pt,
    title=\textbf{Case 4: A Chemistry Item Where Formula Bookkeeping Fails},
    fonttitle=\small\bfseries,
    title after break=\textbf{Case 4: A Chemistry Item Where Formula Bookkeeping Fails (continued)},
    breakable,
    lines before break=6,
    pad at break=3pt
]

\noindent\begin{minipage}{\linewidth}
\noindent\small\textbf{Task.} A Humanity's Last Exam organic-chemistry item asks for the molecular formula of compound \textbf{B}. The prompt describes reactions of tris(2,6-dimethoxyphenyl)methylium ion with either \emph{n}-propylamine or methyl-3-aminopropionate, and the diagram depicts the reaction pathways. Reference answer: \texttt{C23H22NO4+}.

\vspace{4pt}

\begin{center}
\includegraphics[width=0.9\linewidth]{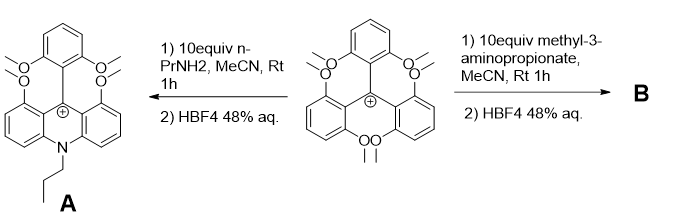}
\end{center}
\end{minipage}

\vspace{4pt}

\noindent\begin{minipage}{\linewidth}
\small\textbf{MiMo-V2.5 \bad{(Incorrect, 0.0)}}

\vspace{2pt}
\begin{tracelist}
(*@\textbf{T1:}@*) read_image("hle_07_image.jpg")
(*@\ret{ -> image parsed; identifies acridinium core and substituents}@*)
(*@\stepp{Reasoning: derives product B from the right-hand pathway.}@*)
(*@\textbf{T2:}@*) write_output("C29H32NO7+")
(*@\bad{Formula does not match the reference C23H22NO4+.}@*)
\end{tracelist}
\end{minipage}

\vspace{6pt}

\noindent\begin{minipage}{\linewidth}
\small\textbf{Qwen3-VL-32B \bad{(Incorrect, 0.0)}}

\vspace{2pt}
\begin{tracelist}
(*@\textbf{T1:}@*) read_image("hle_07_image.jpg")
(*@\ret{ -> describes the reaction diagram and attempts atom accounting}@*)
(*@\stepp{Reasoning: counts atoms after substituent transfer.}@*)
(*@\textbf{T2:}@*) write_output("C31H34NO8")
(*@\bad{Wrong atom count and missing charge; reference is C23H22NO4+.}@*)
\end{tracelist}
\end{minipage}

\end{tcolorbox}

\Needspace{19\baselineskip}
\subsection{Case 5 --- Spatial Plate-Boundary Reasoning}
\label{sec:case5}
\begin{tcolorbox}[
    enhanced, width=\columnwidth,
    colback=blue!3, colframe=blue!40,
    boxsep=1pt, arc=1pt,
    left=3pt, right=3pt, top=3pt, bottom=3pt,
    grow to left by=0pt, grow to right by=0pt,
    before skip=4pt, after skip=6pt,
    title=\textbf{Case 5: A Plate-Map Item Where Option Grounding Fails},
    fonttitle=\small\bfseries,
    title after break=\textbf{Case 5: A Plate-Map Item Where Option Grounding Fails (continued)},
    breakable,
    lines before break=6,
    pad at break=3pt
]

\noindent\begin{minipage}{\linewidth}
\noindent\small\textbf{Task.} An HLE structural-geology item shows a hypothetical tectonic map and asks: ``Along which plate boundary might we expect the longest range of the tallest mountains on the planet shown above? Assume similar tectonic plate geology to Earth.'' The model must choose from nine named plate-boundary options. Reference answer: \texttt{D. South Kesh Plate and Eurybian Plate}.

\vspace{4pt}

\begin{center}
\includegraphics[width=0.9\linewidth]{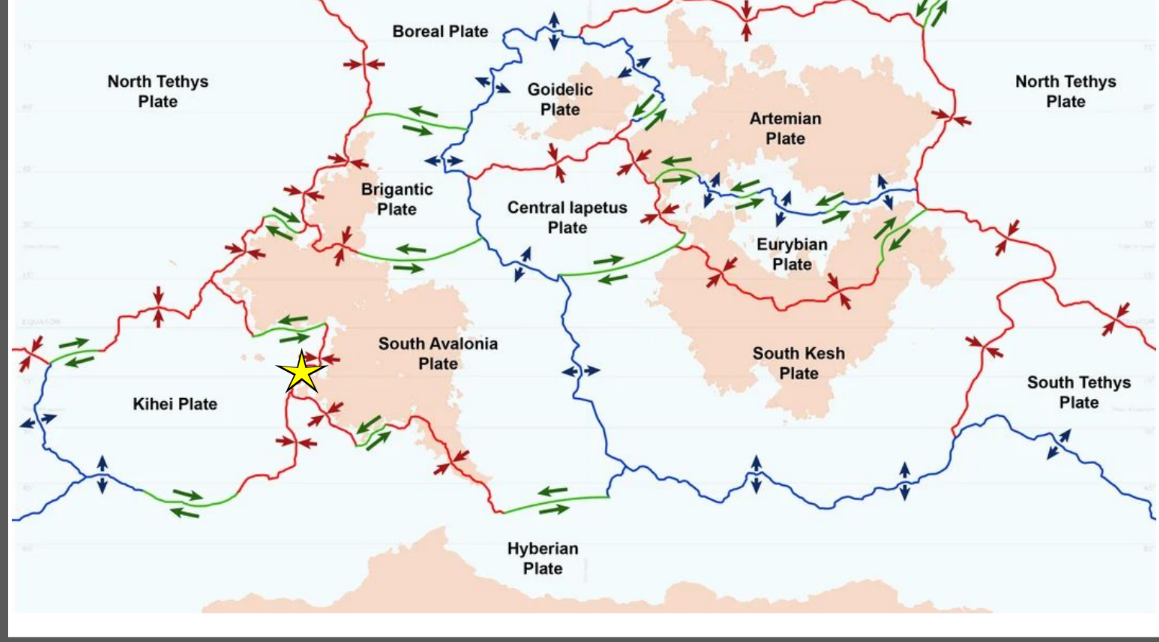}
\end{center}
\end{minipage}

\vspace{4pt}

\noindent\begin{minipage}{\linewidth}
\small\textbf{Claude Opus 4.6 \bad{(Incorrect, 0.0)}}

\vspace{2pt}
\begin{tracelist}
(*@\textbf{T1:}@*) read_image("hle_02_image.jpg")
(*@\ret{ -> reads the map and applies the mountain-building criterion}@*)
(*@\stepp{Reasoning: tallest ranges should form along strong convergent/collision boundaries.}@*)
(*@\textbf{T2:}@*) write_output("G. Artemian Plate and Eurybian Plate")
(*@\bad{Geologic rule is plausible, but the selected option is not the reference boundary D.}@*)
\end{tracelist}
\end{minipage}

\vspace{6pt}

\noindent\begin{minipage}{\linewidth}
\small\textbf{Qwen3.6-35B-A3B \bad{(Incorrect, 0.0)}}

\vspace{2pt}
\begin{tracelist}
(*@\textbf{T1:}@*) read_image("hle_02_image.jpg")
(*@\ret{ -> identifies candidate boundaries and compares plate interactions}@*)
(*@\stepp{Reasoning: searches for the most mountain-building boundary.}@*)
(*@\textbf{T2:}@*) write_output("H")
(*@\bad{Chooses a different boundary option; reference is D, South Kesh Plate and Eurybian Plate.}@*)
\end{tracelist}
\end{minipage}

\end{tcolorbox}

\section{OpenRouter Pricing}
\label{sec:openrouter-pricing}

Table~\ref{tab:openrouter-prices} lists the per-million-token rates used for the USD cost columns in the main paper and Table~\ref{tab:local_cost}. Per-task cost follows the four-term formula $\mathrm{cost}=\tfrac{1}{10^{6}}(u_{\mathrm{in}}\alpha_{\mathrm{in}}+u_{\mathrm{out}}\alpha_{\mathrm{out}}+u_{\mathrm{cr}}\alpha_{\mathrm{cr}}+u_{\mathrm{cw}}\alpha_{\mathrm{cw}})$, where $u_{\mathrm{in}},u_{\mathrm{out}},u_{\mathrm{cr}},u_{\mathrm{cw}}$ are the input, output, cache-read, and cache-write token counts and $\alpha_{\cdot}$ the matching unit prices in Table~\ref{tab:openrouter-prices}; when a model exposes no separate cache-read or cache-write rate (``---''), the corresponding tokens are billed at the input price. All models use OpenRouter listing prices, except for Doubao-Seed2.0-Pro, MiMo-V2-Omni, and MiMo-V2-Pro (marked $\dagger$), which are not listed on OpenRouter and are priced using their official provider rates.

\begin{table*}[!t]
\centering
\small
\setlength{\tabcolsep}{5pt}
\renewcommand{\arraystretch}{1.05}
\begin{tabular}{lllrrrr}
\toprule
\textbf{Model} & \textbf{Provider} & \textbf{Pool} & \textbf{Input} & \textbf{Output} & \textbf{Cache Read} & \textbf{Cache Write}\\
\midrule
\rowcolor{gray!15}
\multicolumn{7}{l}{\textit{API Deployment}} \\
GPT-6-Astra & OpenAI & API & 10.00 & 50.00 & 1.00 & 12.50 \\
GPT-6.1-Sol & OpenAI & API & 2.00 & 10.00 & 0.10 & 2.50 \\
Claude-Opus-5.5 & Anthropic & API & 4.00 & 20.00 & 0.20 & 5.00 \\
Claude-Opus-4.8 & Anthropic & API & 5.00 & 25.00 & 0.50 & 6.25 \\
Claude-Opus-4.6 & Anthropic & API & 5.00 & 25.00 & 0.50 & 6.25 \\
Claude-Sonnet-4.6 & Anthropic & API & 3.00 & 15.00 & 0.30 & 3.75 \\
Claude-Haiku-4.5 & Anthropic & API & 1.00 & 5.00 & 0.10 & 1.25 \\
GPT-5.5 & OpenAI & API & 5.00 & 30.00 & 0.50 & --- \\
GPT-5.4 & OpenAI & API & 2.50 & 15.00 & 0.25 & --- \\
GPT-5.4-mini & OpenAI & API & 0.75 & 4.50 & 0.075 & --- \\
Gemini-3.1-Pro & Google AI Studio & API & 2.00 & 12.00 & 0.20 & 0.375 \\
Gemini-3.5-Flash & Google AI Studio & API & 1.50 & 9.00 & 0.15 & 0.083 \\
Qwen3.8-Max & Alibaba Cloud Int & API & 2.00 & 6.00 & 0.25 & --- \\
Qwen3.6-Plus & Alibaba Cloud Int & API & 0.50 & 3.00 & 0.05 & --- \\
Qwen3.5-Plus & Alibaba Cloud Int & API & 0.40 & 2.40 & 0.04 & --- \\
Doubao-Seed2.0-Pro$^\dagger$ & Volcengine & API & 0.444 & 2.222 & 0.089 & --- \\
Doubao-Seed2.0-Mini & Seed & API & 0.10 & 0.40 & --- & --- \\
GLM-5 & Z.ai & API & 1.00 & 3.20 & 0.20 & --- \\
GLM-4.7 & Z.ai & API & 0.60 & 2.20 & 0.11 & --- \\
Kimi-K3 & Moonshot AI & API & 0.73 & 10.00 & 0.55 & --- \\
Kimi-K2.5 & Moonshot AI & API & 0.60 & 3.00 & 0.10 & --- \\
MiniMax-M2.7 & MiniMax & API & 0.30 & 1.20 & 0.06 & --- \\
DeepSeek-V4-Pro & DeepSeek & API & 0.435 & 0.87 & 0.0036 & --- \\
DeepSeek-V3.2 & Alibaba Cloud Int & API & 0.3705 & 1.112 & 0.0741 & --- \\
MiMo-V2.6-Pro & Xiaomi & API & 0.43 & 0.87 & 0.0036 & --- \\
MiMo-V2.6-Flash & Xiaomi & API & 0.10 & 0.28 & 0.05 & --- \\
MiMo-V2.5 & Xiaomi & API & 0.14 & 0.28 & 0.0028 & --- \\
MiMo-V2.5-Pro & Xiaomi & API & 0.435 & 0.87 & 0.0036 & --- \\
MiMo-V2-Omni$^\dagger$ & Xiaomi & API & 0.14 & 0.28 & 0.0028 & --- \\
MiMo-V2-Pro$^\dagger$ & Xiaomi & API & 0.435 & 0.87 & 0.0036 & --- \\
\midrule
\rowcolor{gray!15}
\multicolumn{7}{l}{\textit{Local Deployment (hosted-reference rates)}} \\
Qwen3.6-27B & Alibaba Cloud Int & Local & 0.60 & 3.60 & --- & --- \\
Qwen3.6-35B-A3B & AkashML & Local & 0.14 & 1.00 & --- & --- \\
Qwen3.5-27B & Alibaba Cloud Int & Local & 0.30 & 2.40 & --- & --- \\
Qwen3.5-35B-A3B & Alibaba Cloud Int & Local & 0.25 & 2.00 & --- & --- \\
Qwen3.5-9B & SiliconFlow & Local & 0.10 & 0.15 & --- & --- \\
Qwen3-Coder-30B-A3B & Alibaba Cloud Int & Local & 0.045 & 2.25 & --- & --- \\
Qwen3-VL-32B & Alibaba Cloud Int & Local & 0.16 & 0.64 & --- & --- \\
Qwen3-VL-8B & Alibaba Cloud Int & Local & 0.18 & 0.70 & --- & --- \\
Gemma-4-26B-A4B & DeepInfra & Local & 0.07 & 0.34 & --- & --- \\
Gemma-3-27B & DeepInfra & Local & 0.08 & 0.16 & --- & --- \\
Qwen2.5-VL-32B & Alibaba Cloud Int & Local & 0.20 & 0.60 & --- & --- \\
Qwen2.5-VL-7B & Alibaba Cloud Int & Local & 0.20 & 0.20 & --- & --- \\
\bottomrule
\end{tabular}
\caption{Per-million-token prices (USD/1M tokens) for all 42 evaluated models. \textbf{Provider} indicates the inference backend from which the rate is sourced. ``---'' denotes that the provider does not expose a separate cache-read or cache-write rate. Entries marked $\dagger$ use the official provider rate (model not listed on OpenRouter).}
\label{tab:openrouter-prices}
\end{table*}

\section{Details of \benchname{}}
\label{sec:datasets}

\subsection{Benchmark Composition}
\label{sec:benchmark-composition}

\begin{figure*}[!t]
\centering
\includegraphics[width=0.82\textwidth]{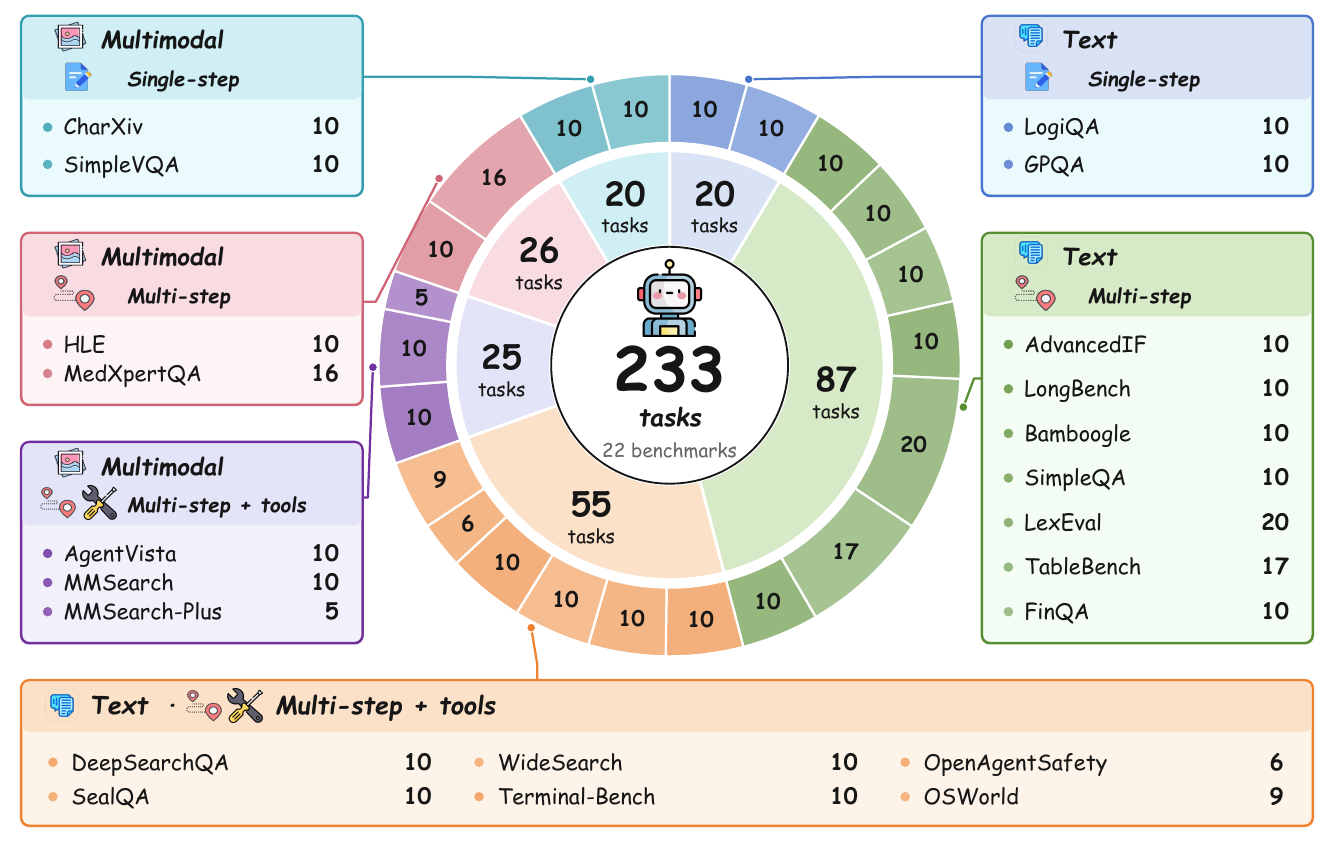}
\caption{Task composition of \benchname{}. The inner ring partitions the 233 tasks along the $2\times3$ modality$\times$execution-complexity taxonomy; the outer ring shows the 22 contributing source benchmarks. Segment sizes are proportional to task counts.}
\label{fig:sunburst}
\end{figure*}

Figure~\ref{fig:sunburst} summarizes the composition of \benchname{}. The suite comprises 233 tasks curated from 22 publicly available source benchmarks through the three-stage (feasibility, difficulty, representativeness) curation procedure described in the main paper. Tasks are partitioned into six workload groups along the input-modality axis~(Text vs.\ Multi-Modal) and the execution-form axis~(Single-Step, Multi-Step, Multi-Step+Tools), yielding group sizes of 20, 87, 55, 20, 26, and 25 for Text-SS, Text-MS, Text-MT, MM-SS, MM-MS, and MM-MT, respectively. The per-benchmark task counts, group assignments, and brief descriptions of each source benchmark are provided in \S\ref{sec:source-benchmarks}.

\subsection{Benchmark Scoring}
\label{sec:scoring-details}

\paragraph{Scoring-mode assignment.}
Each task is assigned exactly one of three scoring modes during benchmark construction, according to the verifiability of its artifact contract: tasks with deterministic answers or artifacts checkable by rule-based procedures use \emph{automated} scoring; tasks with open-ended outputs or semantic constraints that cannot be reliably verified by deterministic rules use \emph{LLM-based} (judge) scoring; and tasks whose contracts contain both rule-verifiable and semantically judged criteria use \emph{hybrid} scoring with fixed per-task weights.

\paragraph{Scoring-mode distribution.}
Table~\ref{tab:scoring_modes} reports the resulting distribution. Overall, 125 tasks~(53.6\%) are scored automatically, 73 tasks~(31.3\%) by the LLM judge, and 35 tasks~(15.0\%) in hybrid mode. The assignment is workload-dependent: multimodal single-step tasks (MM-SS) are exclusively judge-scored because their answers require semantic equivalence assessment against the reference, multimodal tool-use tasks (MM-MT) are exclusively hybrid, and text multi-step tasks (Text-MS, Text-MT) are predominantly automated. All scores produced with judge involvement (judge and hybrid modes) additionally pass through the evidence-based audit described in the main paper and detailed in \S\ref{sec:audit-details}.

\begin{table}[H]
\centering
\small
\setlength{\tabcolsep}{6pt}
\begin{tabular}{lrrrr}
\toprule
\textbf{Group} & \textbf{\#Tasks} & \textbf{Auto} & \textbf{Judge} & \textbf{Hybrid} \\
\midrule
Text-SS  & 20  & 10  & 0  & 10 \\
Text-MS  & 87  & 64  & 23 & 0  \\
Text-MT  & 55  & 35  & 20 & 0  \\
MM-SS & 20  & 0   & 20 & 0  \\
MM-MS & 26  & 16  & 10 & 0  \\
MM-MT & 25  & 0   & 0  & 25 \\
\midrule
\textbf{Total} & \textbf{233} & \textbf{125} & \textbf{73} & \textbf{35} \\
\bottomrule
\end{tabular}
\caption{Distribution of the three scoring modes across the six \benchname{} workload groups. Text/MM = Text/Multi-Modal; SS/MS/MT = Single-Step / Multi-Step / Multi-Step+Tools. Auto = automated (rule-based) scoring; Judge = LLM-based scoring; Hybrid = weighted combination of the two.}
\label{tab:scoring_modes}
\end{table}

\begin{table*}[!t]
\centering
\small
\setlength{\tabcolsep}{4pt}
\renewcommand{\arraystretch}{1.05}
\begin{tabular}{llrll}
\toprule
\textbf{Benchmark} & \textbf{Domain / Capability} & \textbf{\#Tasks} & \textbf{Group} & \textbf{Scoring} \\
\midrule
\rowcolor{gray!15}
\multicolumn{5}{l}{\textit{Text Single-Step (Text-SS)}} \\
LogiQA & Reading comprehension with logical reasoning & 10 & Text-SS & Auto \\
GPQA & Graduate-level science multiple-choice & 10 & Text-SS & Hybrid \\
\midrule
\rowcolor{gray!15}
\multicolumn{5}{l}{\textit{Text Multi-Step (Text-MS)}} \\
AdvancedIF & Complex multi-constraint instruction following & 10 & Text-MS & Judge \\
Bamboogle & Multi-hop question answering & 10 & Text-MS & Auto \\
SimpleQA & Short-form fact-seeking QA & 10 & Text-MS & Auto \\
LongBench & Long-context understanding and retrieval & 10 & Text-MS & Judge \\
LexEval & Legal-domain evaluation & 20 & Text-MS & Auto \\
TableBench & Table understanding and numerical reasoning & 17 & Text-MS & Auto (14) / Judge (3) \\
FinQA & Financial question answering over tables & 10 & Text-MS & Auto \\
\midrule
\rowcolor{gray!15}
\multicolumn{5}{l}{\textit{Text Multi-Step+Tools (Text-MT)}} \\
DeepSearchQA & Deep multi-step search-grounded QA & 10 & Text-MT & Judge \\
WideSearch & Broad-scope information aggregation search & 10 & Text-MT & Auto \\
Seal0 & Search-augmented reasoning on ambiguous queries & 10 & Text-MT & Judge \\
TerminalBench2 & Realistic CLI / shell task execution & 10 & Text-MT & Auto \\
OpenAgentSafety & Open-world agent safety scenarios & 6 & Text-MT & Auto \\
OSWorld & Desktop automation and control & 9 & Text-MT & Auto \\
\midrule
\rowcolor{gray!15}
\multicolumn{5}{l}{\textit{Multi-Modal Single-Step (MM-SS)}} \\
SimpleVQA & Short-form visual fact-seeking QA & 10 & MM-SS & Judge \\
HLE & Humanity's Last Exam (cross-discipline) & 10 & MM-SS & Judge \\
\midrule
\rowcolor{gray!15}
\multicolumn{5}{l}{\textit{Multi-Modal Multi-Step (MM-MS)}} \\
CharXiv & Academic figure and chart understanding & 10 & MM-MS & Judge \\
MedXpertQA & Medical expert-level multimodal QA & 16 & MM-MS & Auto \\
\midrule
\rowcolor{gray!15}
\multicolumn{5}{l}{\textit{Multi-Modal Multi-Step+Tools (MM-MT)}} \\
AgentVista & Ultra-challenging multimodal agent tasks & 10 & MM-MT & Hybrid \\
MMSearch & End-to-end multimodal search & 10 & MM-MT & Hybrid \\
MMSearchPlus & Provenance-aware multimodal search & 5 & MM-MT & Hybrid \\
\midrule
\textbf{Total} & & \textbf{233} & & \\
\bottomrule
\end{tabular}
\caption{Source benchmarks underlying \benchname{}. \#Tasks: number of tasks retained after the three-stage curation. Group: the workload group each benchmark is mapped to under the modality$\times$execution-complexity taxonomy (Text/MM = Text/Multi-Modal; SS/MS/MT = Single-Step / Multi-Step / Multi-Step+Tools). Scoring: the scoring mode(s) used by the benchmark's tasks (Auto = automated, Judge = LLM-based, Hybrid = weighted combination; see \S\ref{sec:scoring-details}).}
\label{tab:dataset_summary}
\end{table*}

\subsection{Source Benchmarks}
\label{sec:source-benchmarks}

\benchname{} curates 233 tasks from 22 widely used open-source benchmarks spanning reasoning, long-context understanding, multi-hop and search-grounded QA, instruction following, tabular and numerical reasoning, vertical domains (finance, law, medicine), code and shell workflows, desktop control, visual question answering, and multimodal search.
Table~\ref{tab:dataset_summary} summarizes each source benchmark, its task count, and the workload group to which it is mapped under our modality$\times$execution-complexity taxonomy~(defined in the main paper's Methodology section).
Brief descriptions of each source benchmark follow.

\noindent \textbullet\ \textbf{LogiQA}~\citep{liu2020logiqa} is a reading-comprehension benchmark for logical reasoning, where each instance provides a short passage and a multiple-choice question whose correct answer requires identifying the logical structure~(premise, conclusion, assumption) of the argument.

\noindent \textbullet\ \textbf{GPQA (Graduate-Level Google-Proof Q\&A)}~\citep{rein2023gpqa} contains Ph.D.-level science questions in biology, chemistry, and physics that are resistant to web search; answering them correctly requires expert-level domain knowledge and multi-step reasoning rather than retrieval.

\noindent \textbullet\ \textbf{AdvancedIF}~\citep{advancedif2025} targets complex instruction following with dozens of verifiable constraints per instruction~(e.g., length, keyword, formatting, and structural requirements), testing a model's ability to satisfy all constraints simultaneously in a single response.

\noindent \textbullet\ \textbf{Bamboogle}~\citep{press2022bamboogle} is a multi-hop question-answering dataset in which each question requires chaining two or more Wikipedia facts; the answer is not directly retrievable and demands cross-document reasoning.

\noindent \textbullet\ \textbf{SimpleQA}~\citep{wei2024simpleqa} is a short-form fact-seeking benchmark where each question has a single, unambiguous answer; it emphasizes precision and resistance to confabulation on entities, dates, and numerical facts.

\noindent \textbullet\ \textbf{LongBench}~\citep{bai2024longbench2} is a long-context benchmark covering summarization, retrieval, and reasoning over documents of up to tens of thousands of tokens; it evaluates a model's ability to locate and synthesize information dispersed across long inputs.

\noindent \textbullet\ \textbf{LexEval}~\citep{li2024lexeval} is a comprehensive legal-domain evaluation suite covering legal knowledge, statute comprehension, and legal reasoning; it tests domain-specific terminology and normative reasoning ability.

\noindent \textbullet\ \textbf{TableBench}~\citep{wu2024tablebench} is a table-understanding benchmark with diverse table structures and question types~(lookup, aggregation, comparison, trend analysis), requiring both precise cell localization and multi-step numerical reasoning.

\noindent \textbullet\ \textbf{FinQA}~\citep{chen2021finqa} is a financial question-answering dataset over earnings reports and financial tables; answering requires programmatic reasoning~(e.g., computing ratios, growth rates) over tabular data accompanied by natural-language context.

\noindent \textbullet\ \textbf{DeepSearchQA}~\citep{deepsearchqa2025} is a deep search benchmark featuring hand-crafted, multi-step information-seeking tasks across 17 fields; unlike single-answer retrieval, it evaluates comprehensiveness and the ability to synthesize evidence from multiple retrieved sources.

\noindent \textbullet\ \textbf{WideSearch}~\citep{widesearch2025} probes broad-scope information aggregation: each task requires gathering and reconciling information from many independent sources, testing the breadth and coverage of a search-augmented agent.

\noindent \textbullet\ \textbf{Seal0}~\citep{sealqa2025} is drawn from the SealQA benchmark for search-augmented language models; its core subset~(Seal-0) contains questions specifically curated to be challenging for frontier models, often involving ambiguity, conflicting evidence, or noisy search results.

\noindent \textbullet\ \textbf{TerminalBench2}~\citep{terminalbench2026} is a command-line interface benchmark in which the agent must accomplish realistic system-administration, data-processing, and development tasks in a Linux terminal, requiring correct sequencing of shell commands and file operations.

\noindent \textbullet\ \textbf{OpenAgentSafety}~\citep{openagentsafety2025} evaluates open-world agent safety through scenarios in which an unconstrained agent may take harmful, reckless, or policy-violating actions; tasks test whether the agent completes the goal while respecting safety constraints.

\noindent \textbullet\ \textbf{OSWorld}~\citep{xie2024osworld} is a desktop-environment benchmark where the agent controls a virtual operating system~(files, applications) to accomplish real-world computer tasks; it demands pixel-level perception, cross-application planning, and long-horizon tool use.

\noindent \textbullet\ \textbf{SimpleVQA}~\citep{visualsimpleqa2025} is a short-form visual fact-seeking benchmark in which each question is grounded in a single image and has one correct answer; it isolates visual entity recognition and fine-grained perception from multi-hop reasoning.

\noindent \textbullet\ \textbf{HLE (Humanity's Last Exam)}~\citep{phan2025hle} is a cross-discipline expert-level benchmark spanning mathematics, humanities, science, and professional domains; questions are designed to be resistant to web search and require deep reasoning, often grounded in images, diagrams, or specialized notation.

\noindent \textbullet\ \textbf{CharXiv}~\citep{wang2024charxiv} is a chart-understanding benchmark built from real arXiv figures; it asks questions that require reading legends, curves, axes, and annotations, testing fine-grained visual reasoning beyond simple OCR or caption matching.

\noindent \textbullet\ \textbf{MedXpertQA}~\citep{zuo2025medxpertqa} is a medical expert-level multimodal QA benchmark covering clinical cases, pathology images, and diagnostic reasoning; it evaluates whether a model can reach expert-level conclusions from multimodal medical evidence.

\noindent \textbullet\ \textbf{AgentVista}~\citep{agentvista2026} is a benchmark for generalist multimodal agents on realistic, ultra-challenging tasks spanning 25 sub-domains; each task requires long-horizon, interleaved tool use grounded in rich visual evidence, testing perception, planning, and tool-use integration jointly.

\noindent \textbullet\ \textbf{MMSearch}~\citep{jiang2024mmsearch} is an end-to-end multimodal search benchmark in which the agent must query, retrieve, and synthesize information that is distributed across both text and images, requiring multimodal query formulation and cross-modal evidence aggregation.

\noindent \textbullet\ \textbf{MMSearchPlus}~\citep{mmsearchplus2025} extends multimodal search with a provenance-aware evaluation: each task demands fine-grained multimodal reasoning while preventing shortcut answers, and correctness is judged by whether the agent cites and correctly interprets the originating evidence.

\begin{figure*}[!b]
\centering
\begin{tcolorbox}[colback=gray!5,colframe=gray!60,title={LLM-as-a-Judge Prompt Template},fonttitle=\bfseries\small,left=3pt,right=3pt,top=3pt,bottom=3pt,boxrule=0.5pt,arc=1pt]
\begin{lstlisting}[basicstyle=\small\ttfamily,breaklines=true,showstringspaces=false,frame=none,framesep=0pt,xleftmargin=0pt,aboveskip=0pt,belowskip=0pt,numbers=none]
You are a grading function. Your ONLY job is to output a single JSON object.

CRITICAL RULES:
- Do NOT use any tools (no Read, Write, exec, or any other tool calls)
- Do NOT create files or run commands
- Do NOT write any prose, explanation, or commentary outside the JSON
- Respond with ONLY a JSON object -- nothing else

Be a strict evaluator. Reserve 1.0 for genuinely excellent performance.
An average acceptable completion should score around 0.6-0.7.
Deduct points for unnecessary steps, verbose output, and inefficient tool usage.

## Task
{task_prompt}

## Expected Behavior
{expected_behavior}

## Agent Transcript (summarized)
{transcript_summary}

## Grading Rubric
{rubric}

Score each criterion from 0.0 to 1.0.

Respond with ONLY this JSON structure (no markdown, no code fences, no extra text):
{"scores": {"criterion_name": 0.0}, "total": 0.0, "notes": "brief justification"}
\end{lstlisting}
\end{tcolorbox}
\caption{LLM-as-a-Judge prompt template. Field placeholders are denoted in curly braces.}
\label{fig:judge-prompt}
\end{figure*}

\newpage
\section{Prompts}
\label{sec:prompts}

\noindent This section documents the two prompt templates used in \benchname{}'s grading pipeline: the LLM-as-a-Judge grading prompt (Figure~\ref{fig:judge-prompt}) and the meta-judge auditing prompt (Figure~\ref{fig:meta-judge-system}). Since all our experiments are conducted under the unified OpenClaw runtime, the system prompt used in the experiments is the default OpenClaw system prompt and is therefore not reported here.

\makeatletter
\setlength{\@dblfptop}{0pt}
\setlength{\@dblfpbot}{0pt plus 1fil}
\makeatother
\begin{figure*}[!t]
\centering
\begin{tcolorbox}[colback=gray!5,colframe=gray!60,title={Meta-Judge System Prompt},fonttitle=\bfseries\small,left=3pt,right=3pt,top=3pt,bottom=3pt,boxrule=0.5pt,arc=1pt]
\begin{lstlisting}[basicstyle=\small\ttfamily,breaklines=true,showstringspaces=false,frame=none,framesep=0pt,xleftmargin=0pt,aboveskip=0pt,belowskip=0pt,numbers=none]
You are a meta-judge auditor for an AI agent benchmark. Your job is to
determine whether a grading judge hallucinated when scoring an agent's work.

You will be given:
1. The task prompt (what the agent was asked to do)
2. The agent's session transcript (what the agent actually did)
3. The judge's score and notes (how the judge graded the agent)
4. Metadata: grading_type, original score, assistant output characters

Your task: Determine if the judge's positive score is justified by the
agent's actual work, or if the judge hallucinated (fabricated evidence
of work that didn't happen).

## Hallucination Patterns (choose one if verdict is "hallucination"):

- rubric-leakage: The judge cites answer text not present in the transcript, apparently copying rubric/ground-truth information.
- context-overflow-confabulation: The agent hit the serving context limit and left an empty/truncated final turn; the judge then fabricated completion.
- pseudo-tool-call-credulity: The agent wrote tool-like text rather than an executable tool call, and the judge credited it as completed.
- workspace-gt-leakage: A workspace-side leakage bug exposed ground-truth or prior-run output and the judge credited it.
- timeout-with-credit: A timed-out trajectory received positive judge credit for unfinished work.
- output-instability: Small-model malformed, unstable, repetitive, or incomplete artifacts were over-interpreted as answers.
- mc-rubric-leak: For multiple-choice tasks, the judge copied the correct option from the rubric instead of the agent trace.
- mc-parse: The judge mis-parsed an MC answer format and credited the wrong choice.
- mc-hedge: The agent hedged or mentioned several MC choices and the judge credited the favorable one.
- mixed: Several taxonomy causes apply and no single one dominates.
- other: A hallucination pattern not covered above.

## Output Format (JSON only, no markdown):
{
  "verdict": "hallucination" | "legitimate",
  "pattern": "<one of the patterns above, or null if legitimate>",
  "confidence": "high" | "medium" | "low",
  "reasoning": "<1-3 sentence explanation>"
}
\end{lstlisting}
\end{tcolorbox}
\caption{Meta-judge system prompt.}
\label{fig:meta-judge-system}
\end{figure*}

\end{document}